\documentclass{article}

\usepackage[preprint]{corl_2022} 

\usepackage{graphicx}

\usepackage{comment}
\usepackage{amsmath,amssymb} 
\usepackage{color}
\usepackage{caption}
\usepackage{subcaption}
\usepackage{dsfont}
\usepackage{booktabs}       
\usepackage{amsfonts}       
\usepackage{nicefrac}       
\usepackage{bbm}
\usepackage{bm}
\usepackage{pifont}
\usepackage{textcomp, gensymb}
\usepackage{fancyvrb}
\usepackage{sidecap}
\usepackage{multirow}
\usepackage{floatrow}
\usepackage{environ}
\usepackage{tabularx}
\usepackage{wrapfig}
\usepackage{parskip}
\usepackage{enumitem}
\newfloatcommand{capbtabbox}{table}[][\FBwidth]
\RecustomVerbatimEnvironment{verbatim}{Verbatim}{fontsize=\tiny}
\usepackage{xspace}
\newcommand*{\eg}{\emph{e.g.}\@\xspace}
\newcommand{\hide}[1]{}

\DeclareRobustCommand\sampleline[1]{%
  \tikz\draw[#1, line width = 1.0] (0,0) (0,\the\dimexpr\fontdimen22\textfont2\relax)
  -- (1em,\the\dimexpr\fontdimen22\textfont2\relax);%
}
\newlength{\savedbdss}
\NewEnviron{multieq}[1]{%
  \[
  \begin{minipage}{\displaywidth}
  \setlength{\savedbdss}{\belowdisplayshortskip}
  \setlength{\abovedisplayshortskip}{0pt}%
  \setlength{\belowdisplayshortskip}{0pt}%
  \begin{tabularx}{\displaywidth}{@{}*{#1}{>{\noindent\vspace*{-\baselineskip}}X}@{}}
  \BODY
  \end{tabularx}
  \vspace{-\belowdisplayskip}
  \vspace{-\savedbdss}
  \end{minipage}
  \]
}

\newcommand{\todo}[1]{{\color{magenta}{\bf\sf [todo: #1]}}}
\newcommand{\name}[0]{Wayformer}

\title{\name{}: Motion Forecasting via Simple \& Efficient Attention Networks }

\author{
  Nigamaa Nayakanti\thanks{Equal contribution.}\\
      \texttt{nigamaa@waymo.com}
  \And Rami Al-Rfou$^*$\\
  \texttt{rmyeid@waymo.com}
  
  \And Aurick Zhou\\
  \texttt{aurickz@waymo.com}
  
  \And Kratarth Goel\\
  \texttt{kratarth@waymo.com}
  
  \And Khaled S. Refaat\\
  \texttt{krefaat@waymo.com}
  
  \And Benjamin Sapp\\
  \texttt{bensapp@waymo.com}
}

\begin{document}
\maketitle


\begin{abstract}
Motion forecasting for autonomous driving is a challenging task because complex driving scenarios result in a heterogeneous mix of static and dynamic inputs.
It is an open problem how best to represent and fuse information about road geometry, lane connectivity, time-varying traffic light state, and  history of a  dynamic set of agents and their interactions into an effective encoding.
To model this diverse set of input features, many approaches proposed to design an equally complex system with a diverse set of modality specific modules.
This results in systems that are difficult to scale, extend, or tune in rigorous ways to trade off quality and efficiency.

In this paper, we present \name{}, a family of attention based architectures for motion forecasting that are simple and homogeneous.
\name{} offers a compact model description consisting of an attention based scene encoder and a decoder.
In the scene encoder we study the choice of early, late and hierarchical fusion of input modalities.
For each fusion type we explore strategies to trade off efficiency and quality via factorized attention or latent query attention. 
We show that early fusion, despite its simplicity of construction, is not only modality agnostic but also achieves state-of-the-art results on both Waymo Open Motion Dataset (WOMD) and Argoverse leaderboards, demonstrating the effectiveness of our design philosophy.
\end{abstract}



\section{Introduction}
In this work, we focus on the general task of future behavior prediction of agents (pedestrians, vehicles, cyclists) in real-world driving environments.
\begin{wrapfigure}{R}{0.5\textwidth}
    \vspace{-1.em}
    \centering
    \includegraphics[width=\textwidth]{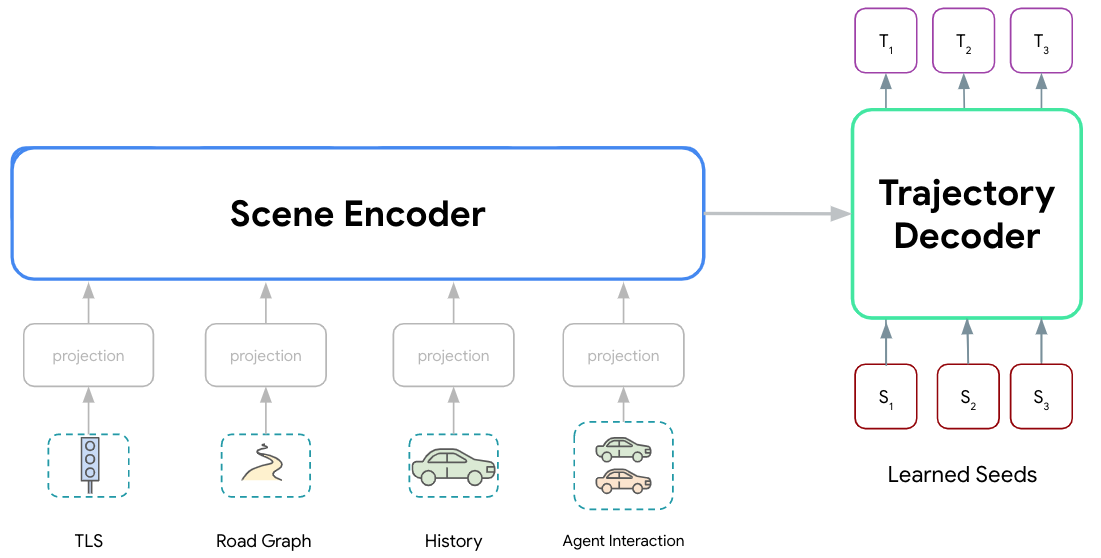}
    \caption{The \name{} architecture as a pair of encoder/decoder Transformer networks. This model takes multimodal scene data as input and produces multimodal distribution of trajectories.}
    \label{fig:wayformer}
\end{wrapfigure}
This is an essential task for safe and comfortable human-robot interactions, enabling high-impact robotics applications like autonomous driving.

The modeling needed for such scene understanding is challenging for many reasons. For one, the \emph{output} is highly unstructured and multimodal---\eg, a person driving a vehicle could carry out one of many underlying intents unknown to an observer, and representing a distribution over diverse and disjoint possible futures is required.
A second challenge is that the \emph{input} consists of a heterogeneous mix of modalities, including agents' past physical state, static road information (\eg location of lanes and their connectivity), and time-varying traffic light information.

Many previous efforts address how to model the multimodal output \cite{rhinehart2018r2p2,multipath,multipathpp,cui2019mtp,tang2019multiple,liang2020garden}, and develop hand-engineered architectures to fuse different input types, each requiring their own preprocessing (\eg, image rasterization \cite{casas2018intentnet,multipath,neural_motion_planner_zeng2019}).
Here, we focus on the multimodality of the \emph{input space}, and develop a simple yet effective modality-agnostic framework that avoids complex and heterogeneous architectures, and leads to a simpler architecture parameterization.
This compact description of a family of architectures results in a simpler design space and allows us to more directly and effectively control for trade-offs in model quality and latency by tuning model computation and capacity.


To keep complexity under control without sacrificing quality or efficiency, we need to find general modeling primitives, which can handle multimodal features that exist in temporal and spatial dimensions concurrently.
Recently, several approaches proposed Transformer networks as the networks of choice for motion forecasting problems \cite{Autobots,multiheadAF,SceneTA,SpatioTemporalGT,agentformer}.
While these approaches offer simplified model architectures, they still require domain expertise and excessive modality specific tuning.
\cite{StackedTransformer} proposed a stack of cross attention layers sequentially processing one modality at a time.
The order in which to process each modality is left to the designer and enumerating all possibilities is combinatorially prohibitive.
\cite{multipathpp} proposed using separate encoders for each modality, where the type of network and its capacity is open for tuning on a per-modality basis. Then modalities' embeddings are flattened and  one single vector is fed to the predictor.
While these approaches allow for many degrees of freedom, they increase the search space significantly.
Without efficient network architecture search or significant human input and hand engineering, the chosen models will likely be sub-optimal given that a limited amount of the modeling options have been explored. 

Our experiments suggest the domain of motion forecasting conforms to Occam's Razor.
We show state of the art results with the simplest design choices and making minimal domain specific assumptions, which is in stark contrast to previous work. When tested in simulation and on real AVs, these Wayformer models showed good understanding of the scene.

Our contributions can be summarized as follows:
\begin{itemize}[noitemsep]
\item
We design a family of models with two basic primitives:  a \emph{self-attention encoder}, where we fuse one or more modalities across temporal and spatial dimensions, and a \emph{cross-attention decoder}, where we attend to driving scene elements to produce a diverse set of trajectories.
\item
We study three variations of the scene encoder that differ in how and when different input modalities are fused.
\item
To keep our proposed models within practical real time constraints of motion forecasting, we study two common techniques to speed up self-attention: \emph{factorized attention} and \emph{latent query attention}.
\item
We achieve state-of-the-art results on both WOMD and Argoverse challenges.
\end{itemize}

\section{Multimodal Scene Understanding}
\label{sec:inputs}
Driving scenarios consist of multimodal data, such as road information, traffic light state, agent history, and agent interactions.
In this section we detail the representation of these modalities in our setup.
For readability, we define the following symbols:
$A$ denotes the number of modeled ego-agents, $T$ denotes the number of past and current timesteps being considered in the history, with a feature size $D_m$.
For a modality $m$, we might have a $4^{th}$ dimension ($S_m$) representing a ``set of contextual objects'' (i.e. representations of other road users) for each modeled agent.

\paragraph{Agent History} contains a sequence of past agent states along with the current state $[A, T, 1, D_h]$.
For each timestep $t \in T$, we consider features that define the state of the agent e.g. x, y, velocity, acceleration, bounding box and so on.  We include a context dimension $S_h = 1$ for homogeneity.

\paragraph{Agent Interactions}
The interaction tensor $[A, T, S_i, D_i]$ represents the relationship between agents.
For each modeled agent $a \in A$, a fixed number of the closest context agents $c_i \in S_i$ around the modeled agent are considered.  
These context agents represent the agents which influence the behavior of our modeled agent.
The features in $D_i$ represent the physical state of each context agents (as in $D_h$ above), but transformed into the frame of reference of our ego-agent.

\paragraph{Roadgraph}
The roadgraph $[A, 1, S_r, D_r]$ contains road features around the agent.
Following \cite{multipath}, we represent roadgraph segments as polylines, approximating the road shape with collections of line segments specified by their endpoints and annotated with type information.
We use $S_r$ roadgraph segments closest to the modeled agent.
Note that there is no time dimension for the road features, but we include a time dimension of 1 for homogeneity with the other modalities.

\paragraph{Traffic Light State}
For each agent $a \in A$, traffic light information $[A, T, S_{tls}, D_{tls}]$ contains the states of the traffic signals that are closest to that agent. 
Each traffic signal point $tls \in S_{tls}$ has features $D_{tls}$ describing the position and confidence of the signal.

\section{\name{}}
\label{sec:network}
We design the family of \name{} models to consist of two main components: a Scene Encoder and a Decoder.
The scene encoder is mainly composed of one or more attention encoders that summarize the driving scene. 
The decoder is a stack of one or more standard transformer cross-attention blocks, in which learned initial queries are fed in, and then cross-attended with the scene encoding to produce trajectories. Figure \ref{fig:wayformer} shows the \name{} model processing  multimodal inputs to produce scene encoding.
This scene encoding serves as the context for the decoder to generate $k$ possible trajectories covering the multimodality of the output space.

\paragraph{Frame of Reference}
As our model is trained to produce futures for a single agent, we  transform the scene into an ego-centric frame of reference by centering and rotating the scene's spatial features around the ego-agent's position and heading at the current time step. 

\paragraph{Projection Layers}
Different input modalities may not share the same number of features, so we project them to a common dimension $D$ before concatenating all modalities along the temporal and spatial dimensions [$S$, $T$].
We found the simple transformation $\text{ Projection}(x_i) = \mathrm{relu}(\mathbf{W}x_i+b)$, where $x_i \in \mathbb{R}^{D_m}$, $b \in \mathbb{R}^{D}$, and $\mathbf{W} \in \mathbb{R}^{D\times D_m }$, to be sufficient.
Concretely, given an input of shape $[A, T, S_m, D_m]$ we project its last dimension producing a tensor of size $[A, T, S_m, D]$.
\begin{figure}[t]
\centering
\begin{subfigure}[b]{0.325\columnwidth}
\includegraphics[scale=0.24, trim={0em 1em 107em 6em},clip]{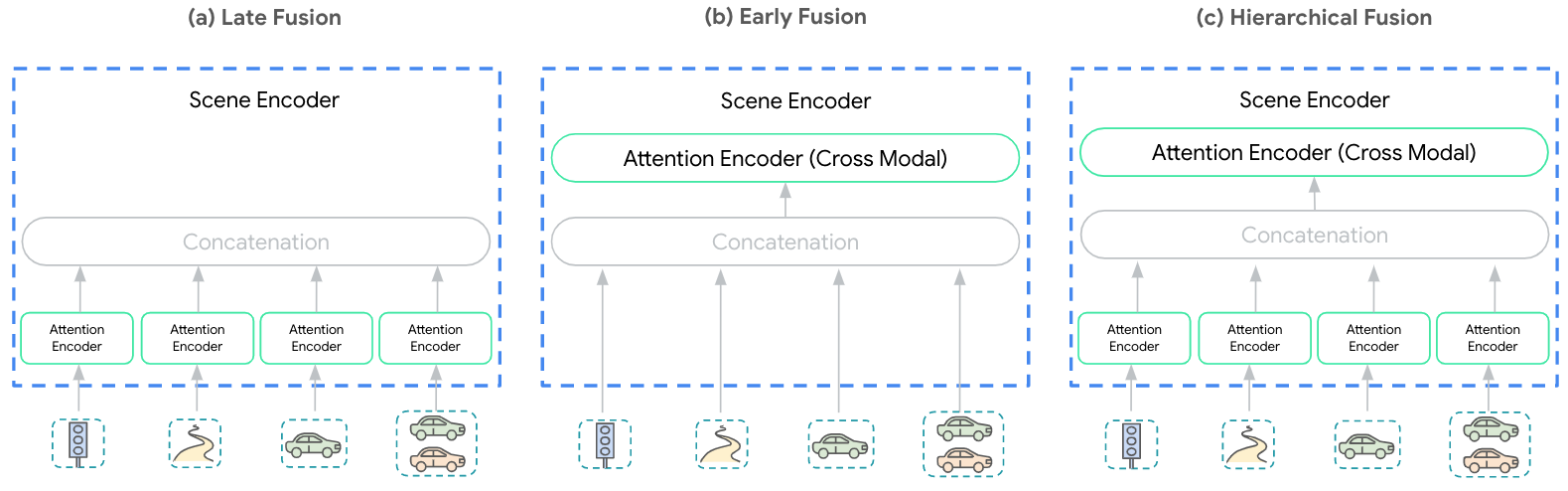}
\caption{Late Fusion}
\label{fig:latefusion_encoder}
\end{subfigure}
\begin{subfigure}[b]{0.325\columnwidth}
\includegraphics[scale=0.24, trim={54em 1em 54em 6em},clip]{figures/figure2.png}
\caption{Early Fusion}
\label{fig:earlyfusion_encoder}
\end{subfigure}
\begin{subfigure}[b]{0.325\columnwidth}
\includegraphics[scale=0.24, trim={107em 1em 0em 6em},clip]{figures/figure2.png}
\caption{Hierarchical Fusion}
\label{fig:hierarchicalfusion_encoder}
\end{subfigure}
\caption{\name{} scene encoder fusing multimodal inputs at different stages. Late fusion dedicates an attention encoder per modality while early fusion process all inputs within one cross modal encoder. Finally, hierarchical fusion combines both the approaches.}
\label{fig:fusion}
\end{figure}
\paragraph{Positional Embeddings}
Self-attention is naturally permutation equivariant, therefore, we may think of them as set-encoders rather than sequence encoders.
However, for modalities where the data does follow a specific ordering, for example agent state across different time steps, it is beneficial to break permutation equivariance and utilize the sequence information.
This is commonly done through positional embeddings.
For simplicity, we add learned positional embeddings for all modalities.
As not all modalities are ordered, the learned positional embeddings are initially set to zero, letting the model learn if it is necessary to utilize the ordering within a modality.
\subsection{Fusion}
\label{sec:fusion}
Once projections and positional embeddings are applied to different modalities, the scene encoder combines the information from all modalities to generate a representation of the environment.
Concretely, we aim to learn  a scene representation $\boldsymbol{Z} = Encoder(\{m_0, m_1, ..., m_k\}),$ where $m_i \in \mathbb{R}^{A \times (T \times S_m) \times D}$, $\boldsymbol{Z} \in \mathbb{R}^{A \times L \times D}$, and $L$ is a hyperparameter.

However, the diversity of input sources makes this integration a non-trivial task.
Modalities might not be represented at the same abstraction level or scale: \{pixels vs objects\}.
Therefore, some modalities might require more computation than the others.
Splitting compute and parameter count among modalities is application specific and non-trivial to hand-engineer.
We attempt to simplify the process by proposing three levels of fusion: \{Late, Early, Hierarchical\}.

\paragraph{Late Fusion}
This is the most common approach used by motion forecasting models, where each modality has its own dedicated encoder (See Figure \ref{fig:fusion}).
We set the width of these encoders to be equal to avoid introducing extra projection layers to their outputs.
Moreover, we share the same depth across all encoders to narrow down the exploration space to a manageable scope.
Transfer of information across modalities is allowed only in the cross-attention layers of the trajectory decoder.

\paragraph{Early Fusion}
Instead of dedicating a self-attention encoder to each modality, early fusion reduces modality specific parameters to only the projection layers (See Figure \ref{fig:fusion}).
In this paradigm, the scene encoder consists of a single self-attention encoder (``Cross-Modal Encoder"), giving the network maximum flexibility in assigning importance across modalities with minimal inductive bias.
\paragraph{Hierarchical Fusion}
As a compromise between the two previous extremes, capacity is split between modality-specific self-attention encoders and the cross-modal encoder in a hierarchical fashion.
As done in late fusion, width and depth is common across  attention encoders and the cross modal encoder.
This effectively splits the depth of the scene encoder between modality specific encoders and the cross modal encoder (Figure \ref{fig:fusion}).
\begin{figure}[t]
  \begin{subfigure}[b]{0.48\columnwidth}
    \includegraphics[height=0.55\linewidth,width=\linewidth]{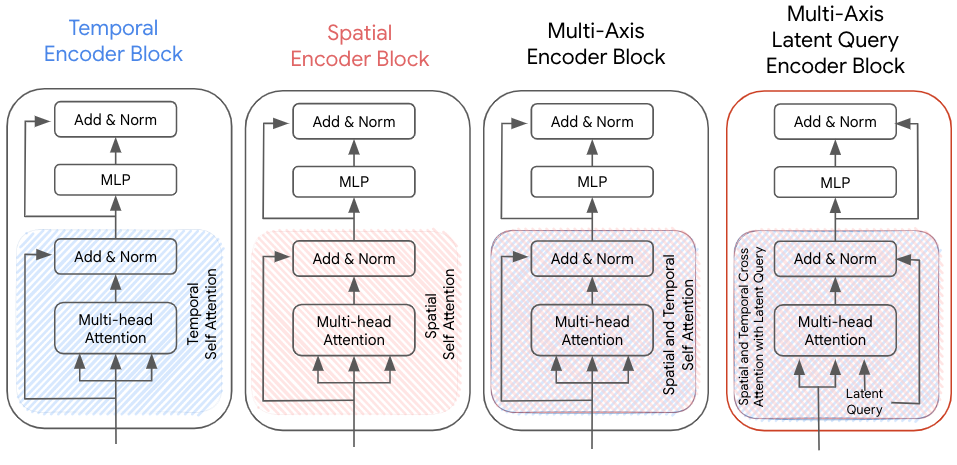}
    \caption{Encoder Blocks}
    \label{fig:encoder-blocks}
  \end{subfigure}
  \begin{subfigure}[b]{0.48\columnwidth}
    \includegraphics[height=0.55\linewidth,width=\linewidth]{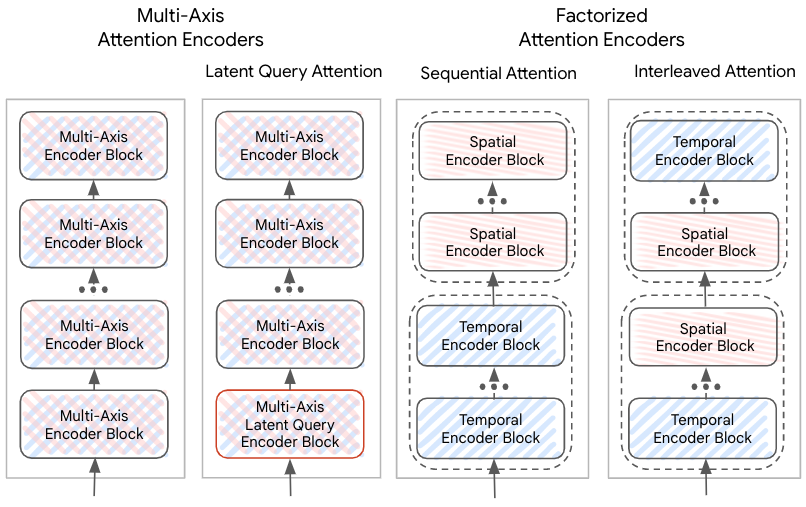}
    \caption{Encoders}
    \label{fig:encoders}
  \end{subfigure}
  \caption{A summary of encoder architectures considered for Wayformer. (a) provides an overview of different encoder blocks and (b) explains how these blocks are arranged to construct the encoder. \vspace{-2em}}
  \label{fig:blocks-encoders}
\end{figure}
\subsection{Attention}
\label{sec:partial}
Transformer networks do not scale well for large multidimensional sequences due to two factors: (a) Self-attention is quadratic in  the input sequence length.  (b) Position-wise Feed-forward networks are expensive sub-networks.
In the following sections, we discuss different speedups to the transformer networks that will help us scale more effectively.
\paragraph{Multi-Axis Attention}
This refers to the default transformer setting which applies self-attention across both spatial and temporal dimensions simultaneously (See Figure ~\ref{fig:encoders}), which we expect to be the most expensive computationally. Computational complexity of early, late and hierarchical fusions with multi-axis attention is $\mathcal{O}(S_m^2 \times T^2)$.

\paragraph{Factorized Attention}
Computational complexity of the self-attention is a quadratic in input sequence length. This becomes more pronounced in multi-dimensional sequences, since each extra dimension increases the size of the input by a multiplicative factor. For example, some input modalities have both temporal and spatial dimensions, so the compute cost scales as $\mathcal{O}(S_m^2 \times T^2)$. To alleviate this, we consider factorized attention \cite{vivit,axialattn} along the two dimensions.
This exploits the  multidimensional structure of input sequences by applying self-attention over each dimension individually, which reduces the cost of self-attention sub-network from $\mathcal{O}(S_m^2\times T^2)$ to $\mathcal{O}(S_m^2)+\mathcal{O}(T^2)$.
Note that the linear term still tends to dominate if $\sum_m \mathcal{S}_m\times T << 12 \times D$ \cite{scalinglaws1}. 

While factorized attention has the potential to reduce computation compared to multi-axis attention, it introduces  complexity in deciding the order in which self-attention is applied to each dimension.
In our work, we compare two paradigms of factorized attention (see Figure~\ref{fig:encoders}):

\begin{itemize}[noitemsep]
    \item \textbf{Sequential Attention:} an $N$ layer encoder consists of $N/2$ temporal encoder blocks followed by another $N/2$ spatial encoder blocks.
    \item \textbf{Interleaved Attention:} an $N$ layer encoder consists of temporal and spatial encoder blocks alternating $N/2$ times.
\end{itemize}

\paragraph{Latent Query Attention}
\label{sec:latents}
Another approach to address the computational costs of large input sequences is to use latent queries \cite{settransformer,perceiver} in the first encoder block, where input $x \in \mathbb{R}^{A \times L_{\text{in}} \times D}$ is mapped to latent space $z \in \mathbb{R}^{A \times L_{\text{out}} \times D}$.
These latents $z \in \mathbb{R}^{A \times L_{\text{out}} \times D}$ are processed further by a series of encoder blocks that take in and return arrays in this latent space (see Figure \ref{fig:encoder-blocks}).
This gives us full freedom to set the latent space resolution, reducing the computational costs of the both self-attention component and the position-wise feedforward network of each block. 
We set the reduction value $\left(R = L_{\text{out}}/L_{\text{in}}\right)$ to be a percentage of the input sequence length. Reduction factor $R$ is kept constant across all the attention encoders in late and hierarchical fusions.

\subsection{Trajectory Decoding}
As our focus is on how to integrate information from different modalities in the encoder, we simply follow the training and output format of \cite{multipath,multipathpp}, where the \name{} predictor outputs a mixture of Gaussians to represent the possible trajectories an agent may take.
To generate predictions, we use a Transformer decoder which is fed a set of $k$ learned initial queries ($\boldsymbol{S}_i \in \mathbb{R}^{h})_{i=1}^k$ and cross attends them with the scene embeddings from the encoder in order to generate embeddings for each component in the output mixture of Gaussians. 

Given the embedding $Y_i$ for a particular component of the mixture, we estimate the mixture likelihood with a linear projection layer that produces the unnormalized log-likelihood for the component.
To generate the trajectory, we project $Y_i$ using another linear layer to output 4 time series: $T_i =  \{\mu_x^t, \mu_y^t, \log \sigma_x^t, \log \sigma_y^t\}_{t=1}^T$ corresponding to the means and log-standard deviations of the predicted Gaussian at each timestep. 

\hide{
\begin{multieq}{2}
\begin{equation}
    score(Y_i) = \mathbf{W}_1Y_i + b_1 \in \mathbb{R}
\end{equation}
&
\begin{equation}
    \Pr(i | \boldsymbol{Y}) = \frac{e^{\nicefrac{score(Y_i)}{\tau}}}{\sum_{j=1}^m e^{\nicefrac{score(Y_j)}{\tau}}}
\end{equation}
&
\begin{equation}
    T_i =  \mathbf{W}_2Y_i + b_2 \in \mathbb{R}^{t*4}
\end{equation}
&
\scalebox{0.875}{
\begin{equation}
    \Pr(G \mid \boldsymbol{X}) =  \sum_{j=1}^m \Pr(j | \boldsymbol{Y}) \prod_{t=1}^l  \phi(G_t | \boldsymbol{S}_j, \boldsymbol{X})
\end{equation}
}
&
\begin{equation}
    \phi(G_t | \boldsymbol{S}_j, \boldsymbol{X}) = \mathcal{N}(G_t | \mu_x^j, \mu_y^j, \sigma_x^j, \sigma_y^j)
\end{equation}
\end{multieq}
}

During training, we follow \cite{multipath,multipathpp} in decomposing the loss into separate classification and regression losses. 
Given $k$ predicted Gaussians $(T_i)_{i=1}^k$, let $\hat i$ denote the index of the Gaussian with mean closest to the ground truth trajectory $G$. 
We train the mixture likelihoods on the log likelihood of selecting the index $\hat i$, and the Gaussian $T_{\hat i}$ to maximize the log-probability of the ground truth trajectory.
\begin{align}
\label{eq:obj}
    \max \underbrace{\log \Pr(\hat i \mid Y)}_{\text{classification loss}} + \underbrace{\log \Pr(G | T_{\hat i})}_{\text{regression loss}}.
\end{align}
\vspace{-2em}

\subsection{Trajectory Aggregation} \label{subsec:nms}
If the predictor outputs a GMM with many modes, it can be difficult to reason about a mixture with so many components, and the benchmark metrics often restrict the number of trajectories being considered.
During evaluation, we thus apply trajectory aggregation following \cite{multipathpp} in order to reduce the number of modes being considered while still preserving the diversity in the original output mixture. 
We refer the reader to Appendix \ref{app:nms} and \cite{multipathpp} for details of the aggregation scheme.

\section{Experimental Setup}

\subsection{Datasets}
\label{sec:datasets}
\paragraph{Waymo Open Motion Dataset (WOMD)} consists of 1.1M examples time-windowed from 103K 20s scenarios derived from real-world driving in urban and suburban environments.
Each example consists of 1 second of history state and 8 seconds of future, which we resample at 5Hz.
The object-agent state contains attributes such as position, agent dimensions, velocity and acceleration vectors, orientation, angular
velocity, and turn signal state.
The long (8s) time horizon in this dataset tests the model’s ability to capture a large field of view and scale to a large output space of trajectories.
\paragraph{Argoverse Dataset} consists of 333K scenarios containing trajectory histories, context agents, and lane centerline inputs for motion prediction.
The trajectories are sampled at 10Hz, with 2 seconds of history and a 3-second future prediction horizon.

\subsection{Training Details and Hyperparameters}

We compare models using competition specific metrics associated with these datasets (see Appendix~\ref{app:metrics}). 
For all metrics, we consider only the top $k=6$ most likely modes output by our model (after trajectory aggregation) and use only the mean of each mode.

For all experiments, we train models using the AdamW optimizer \cite{adamw} with an initial learning rate of 2e-4 and linearly decaying to 0 over 1M steps.
We train models using 16 TPU v3 cores each, with a batch size of 16 per core, resulting in a total batch size of 256 examples per step. 

To vary the capacity of the models, we consider hidden sizes among $\{64, 128, 256\}$ and depths among $\{1, 2, 4\}$ layers.
We fix the intermediate size in the feedforward network of the Transformer block to be either 2 or 4 times the hidden size.

\begin{wrapfigure}{R}{0.4\textwidth}
\vspace{-0.15em}
\vspace*{\fill}
\centering
\vspace*{\fill}
\vspace{-10pt}
{\includegraphics[height=0.5\textwidth,width=\textwidth]{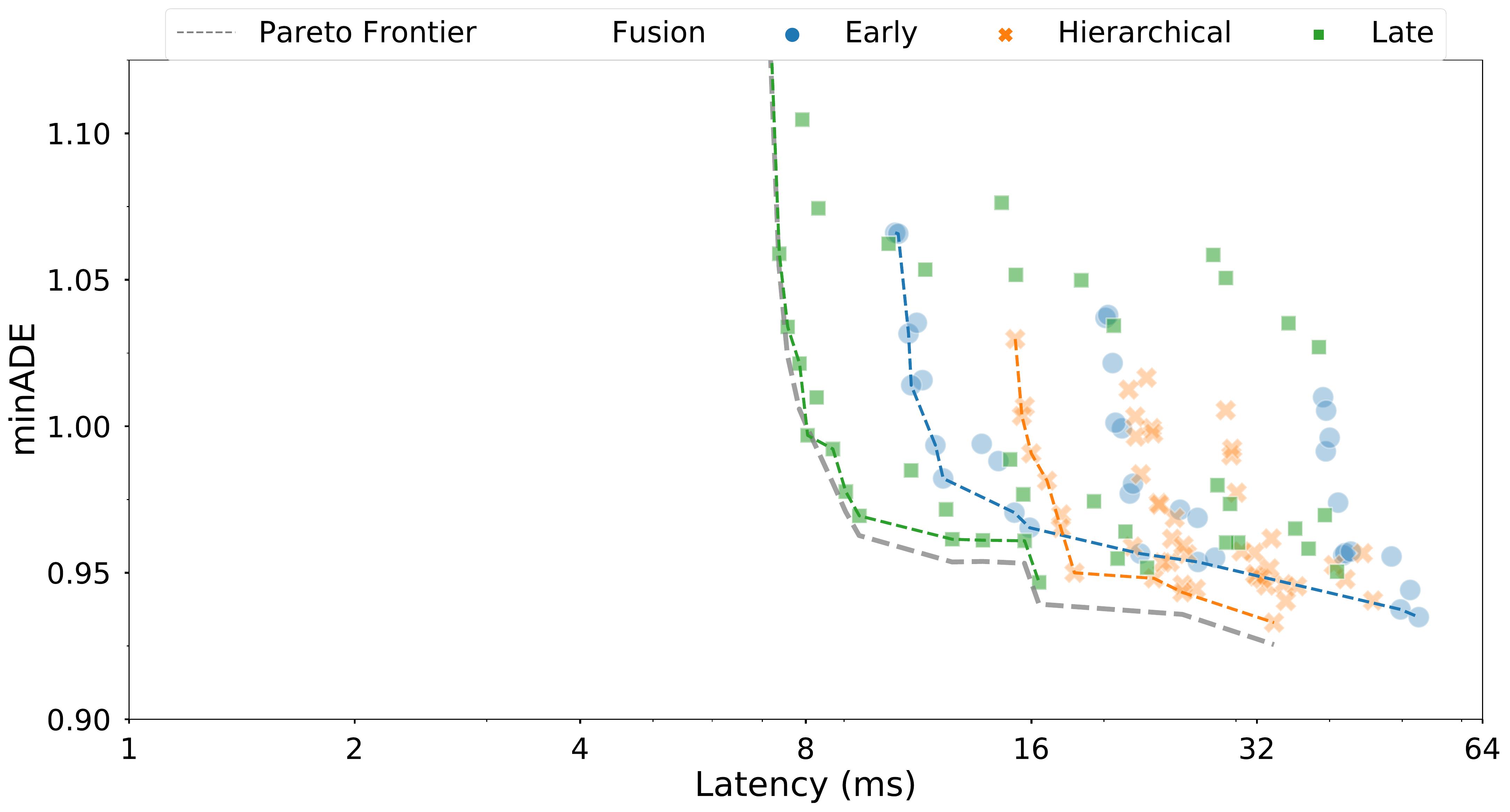}
\subcaption{Efficiency}
\label{fig:fusion-result-latency}}
 \par
 \vfill
 \vspace{1em}
{\includegraphics[height=0.5\textwidth,width=\textwidth]{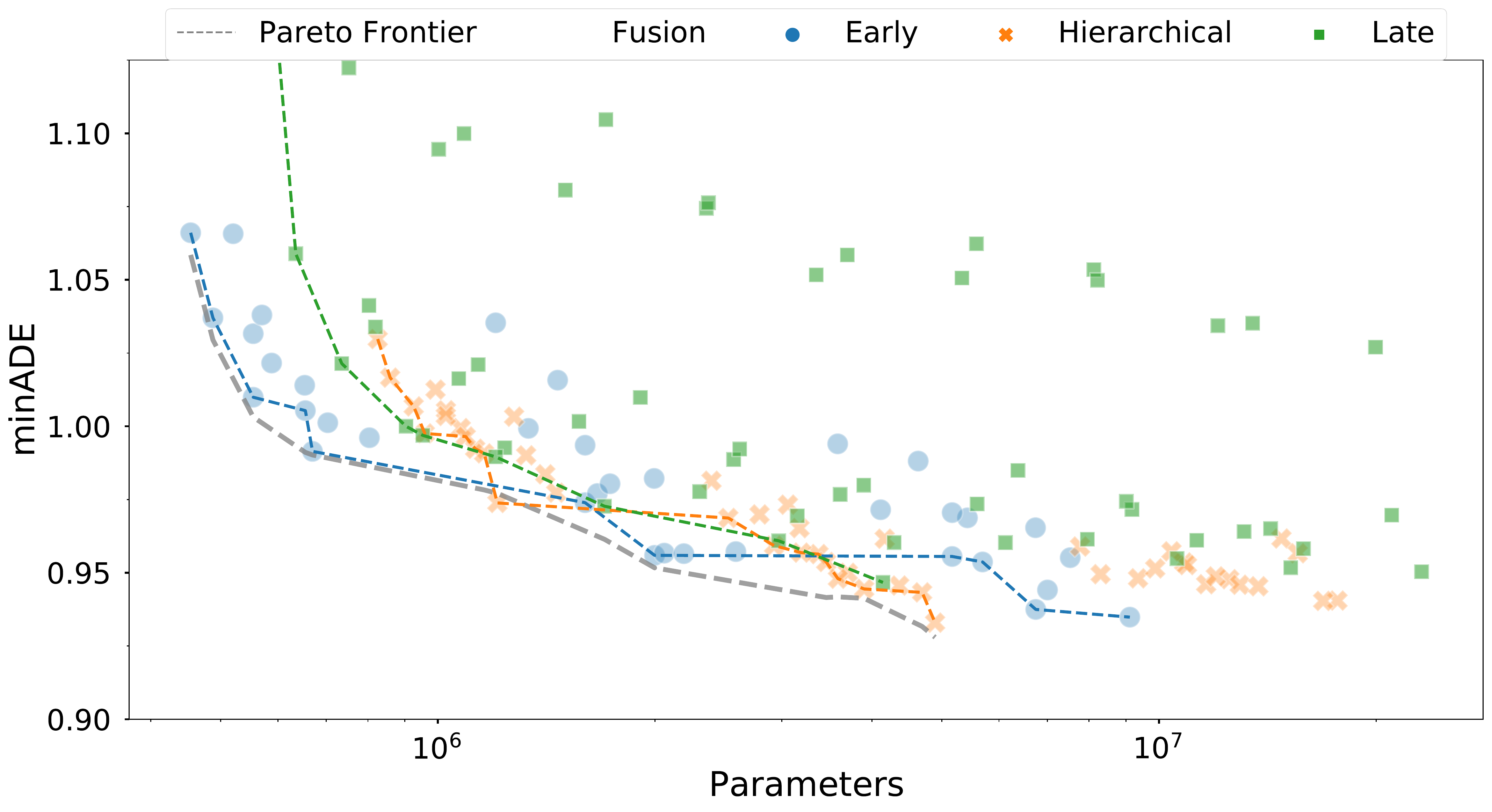}
\subcaption{Capacity}
\label{fig:fusion-result-params}}
\caption{\vspace{-2pt} MinADE of different fusion models with multi-axis attention.\vspace{-25pt}}
\label{fig:fusion-results}
\end{wrapfigure}


For our architecture study in Sections (\ref{sec:fusion-results}-\ref{sec:results-latents}), each predictor outputs a mixture of Gaussians with $m=6$ components, with no trajectory aggregation.
For our benchmark results in Section \ref{sec:sota}, each predictor outputs a mixture of Gaussians with $m=64$ components, and we prune the mixture components using the trajectory aggregation scheme described in Section \ref{subsec:nms}.
For experiments with latent queries, we experiment with reducing the original input resolution to $0.25, 0.5$, $0.75$ and $0.9$ times the original sequence length.
We include a full description of hyperparameters in Appendix \ref{app:hyper}.

\section{Results}
\label{sec:results}

In this Section, we present experiments that demonstrate the trade-offs of combining different fusion strategies with vanilla self-attention (multi-axis) and more optimized methods such as factorized attention and learned queries.
In our ablation studies (Section \ref{sec:fusion-results}-\ref{sec:results-latents}),
we trained models with varying capacities (0.3M-20M parameters) for 1M steps on WOMD.
We report their inference latency on a current generation GPU, capacity, and minADE as a proxy of quality.

\subsection{Multi-Axis Attention}
\label{sec:fusion-results}
In these experiments, we train Wayformer models 
on early, hierarchical and late fusion (Section \ref{sec:fusion}) in combination with multi-axis attention.
In Figure (\ref{fig:fusion-result-latency}),
we show that for models with low latency ($x \leq 16$ ms), late fusion represents an optimal choice.
These models are computationally cheap since there is no interaction between modalities during the scene encoding step.
Adding the cross modal encoder for hierarchical models unlocks further quality gains for models in the range
($16$ms $< x < 32$ms).
Finally, we can see that early fusion can match hierarchical fusion at higher computational cost ($x > 32$ms).
We then study the model quality as a function of capacity, as measured by the number of trainable parameters (Figure \ref{fig:fusion-result-params}).
Small models perform best with early fusion, but as model capacity increases, sensitivity to the choice of fusion decreases dramatically.

\subsection{Factorized Attention}
To reduce the computational budget of our models, we train models with factorized attention instead of jointly attending to spatial and temporal dimensions together.
When combining different modalities together for the cross modal encoder, we first tile the roadgraph modality to a common temporal dimension as the other modalities, then concatenate modalities along the spatial dimension.
After the scene encoder, we pool the encodings over the time dimension before feeding to the predictor.

\begin{figure}[h]
  \begin{subfigure}[b]{0.32\columnwidth}
    \includegraphics[width=\linewidth]{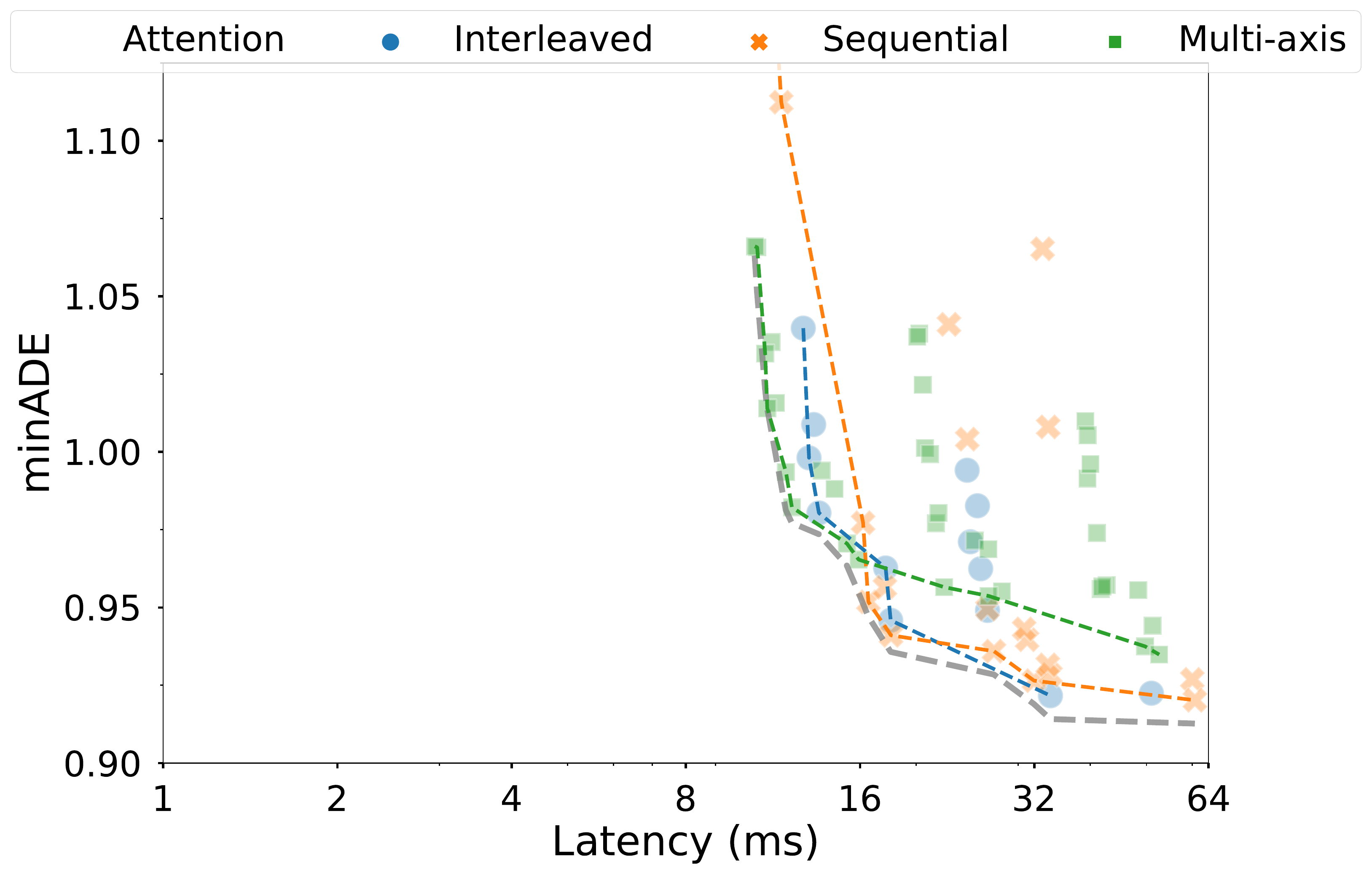}
    \caption{Early Fusion.}
    \label{fig:early-axis-results}
  \end{subfigure}
  \begin{subfigure}[b]{0.32\columnwidth}
    \includegraphics[width=\linewidth]{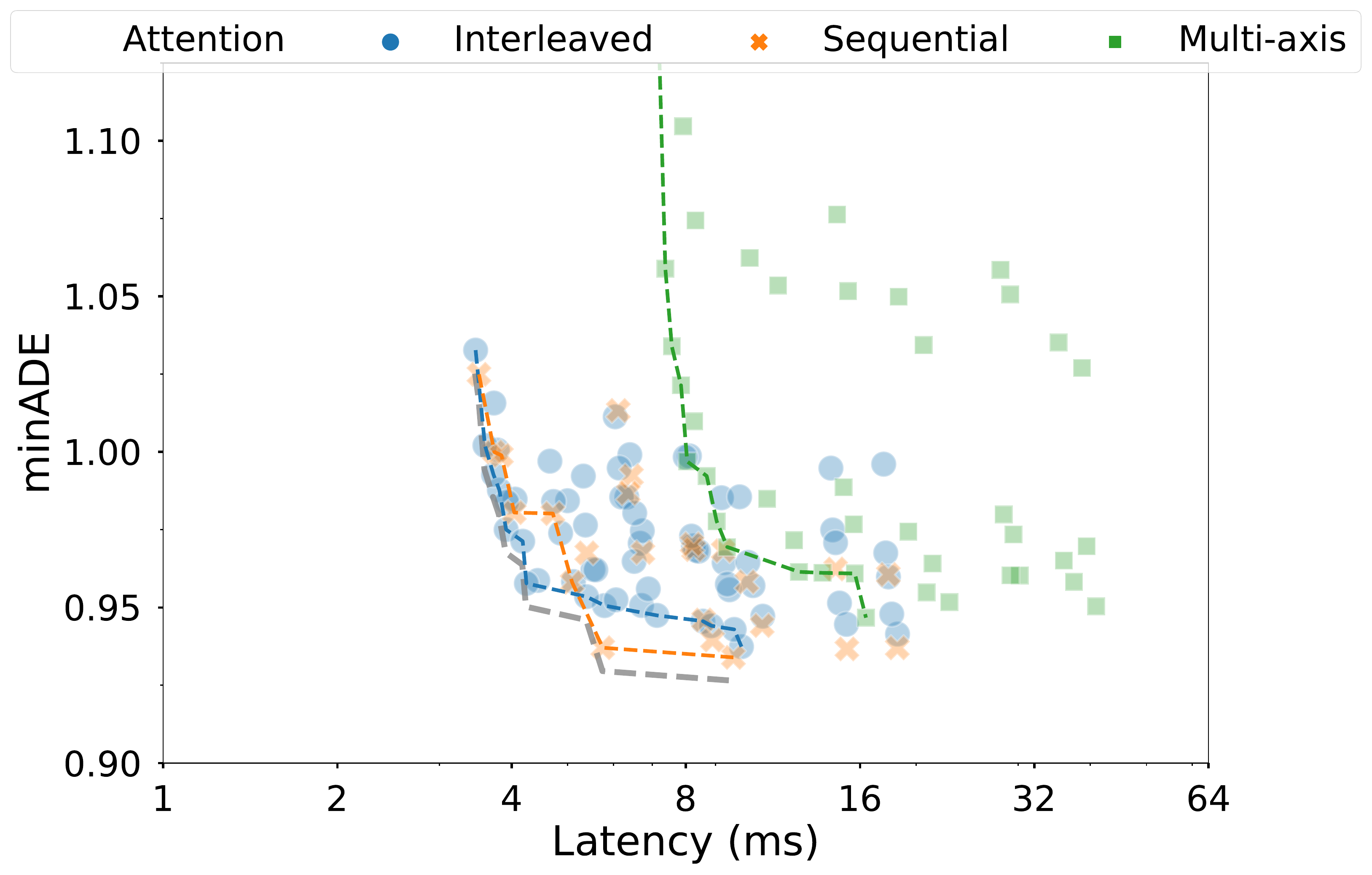}
    \caption{Late Fusion.}
    \label{fig:late-axis-results}
  \end{subfigure}
  \begin{subfigure}[b]{0.32\columnwidth}
    \includegraphics[width=\linewidth]{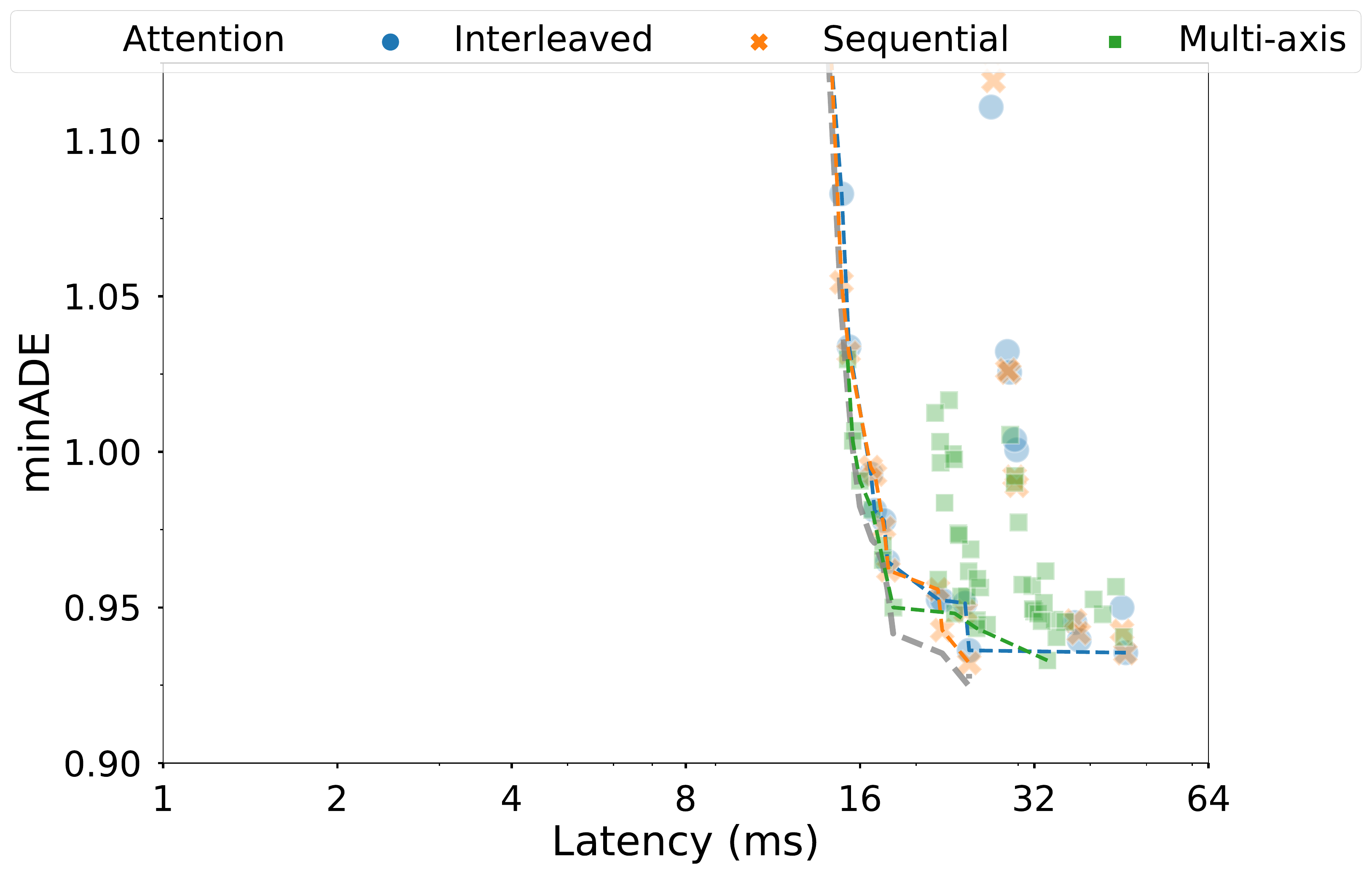}
    \caption{Hierarchical Fusion.}
    \label{fig:hier-axis-results}
  \end{subfigure}
  \caption{Factorized attention improves quality, but only speeds up late fusion models.}
  \label{fig:axis-att}
\end{figure}

We study two types of factorized attention: {sequential, interleaved} (Figure \ref{fig:axis-att}).
First, we observe that both sequential and interleaved factorized attention perform similarly across all types of fusion.
Second, we are surprised to see quality gains from applying factorized attention to the early and late fusion cases (Figures \ref{fig:early-axis-results}, \ref{fig:late-axis-results}).
Finally, we only observe latency improvements for late fusion models (Figure \ref{fig:late-axis-results}), since tiling the road graph to the common temporal dimension in cross-modal encoder used in early and hierarchical fusion significantly increases the count of tokens.

\subsection{Latent Queries}
\label{sec:results-latents}
In this study, we train models with multi-axis latent query encoders with varying levels of input sequence length reduction in the first layer as shown in Figure \ref{fig:axis-att}.
The number of the latent queries is calculated to be a percentage of the input size of the Transformer network with $0.0\%$ indicating the baseline models (multi-axis attention with no latent queries as presented in Figure \ref{fig:fusion-results}).

\begin{figure}[h]

  \begin{subfigure}[b]{0.32\columnwidth}
    \includegraphics[width=\linewidth]{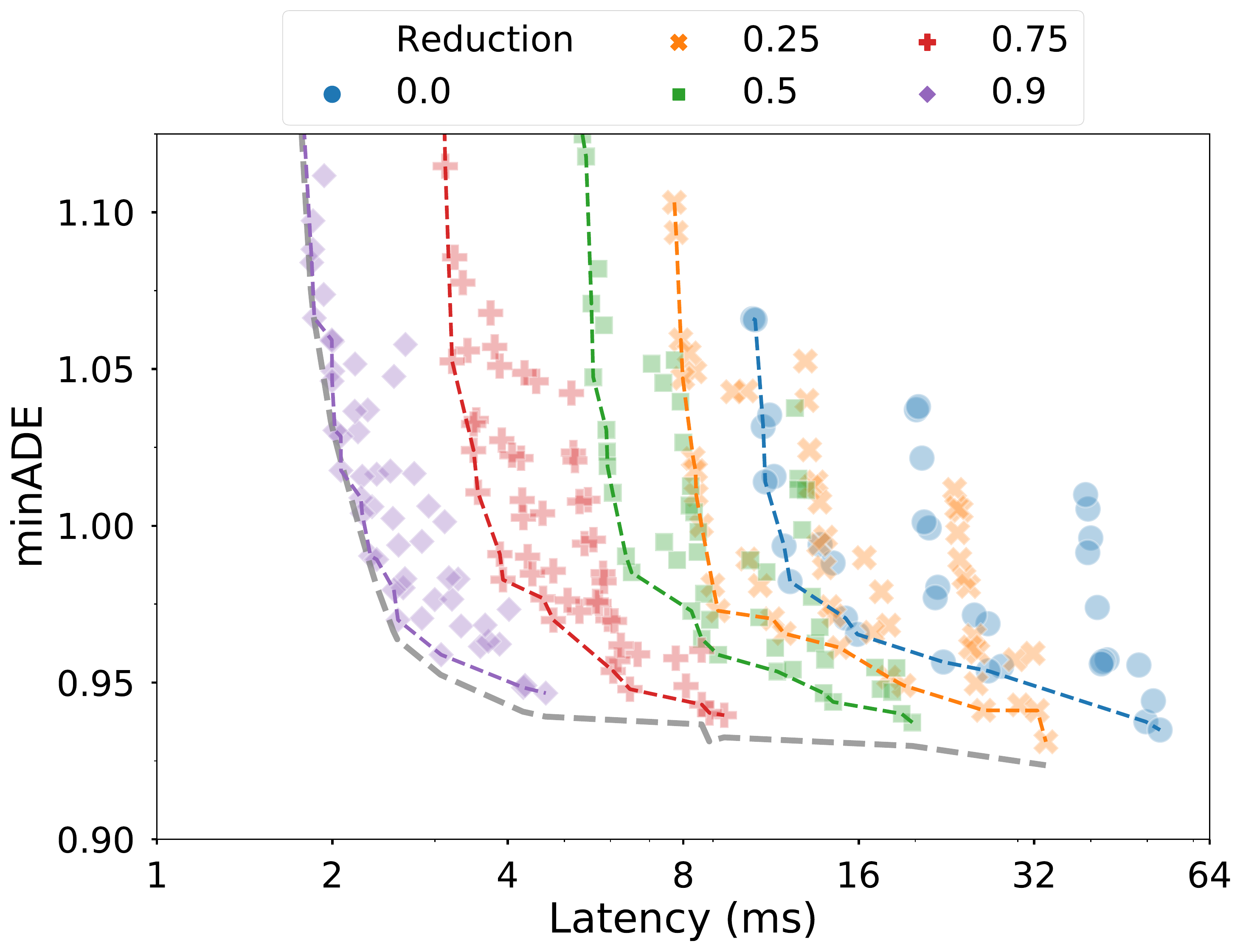}
    \caption{Early Fusion.}
    \label{fig:early-lq-results}
  \end{subfigure}
  \begin{subfigure}[b]{0.32\columnwidth}
    \includegraphics[width=\linewidth]{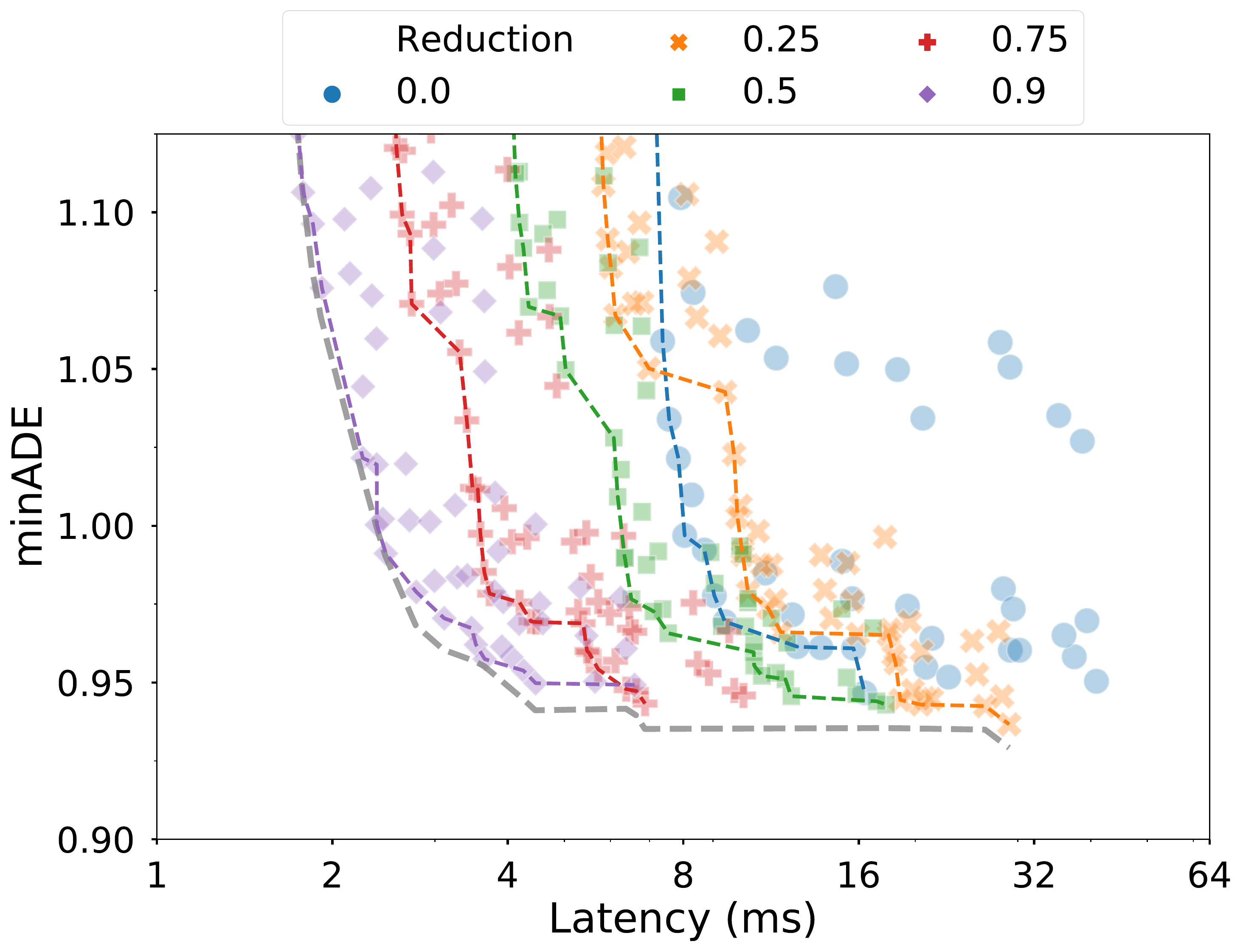}
    \caption{Late Fusion.}
    \label{fig:late-lq-results}
  \end{subfigure}
  \begin{subfigure}[b]{0.32\columnwidth}
    \includegraphics[width=\linewidth]{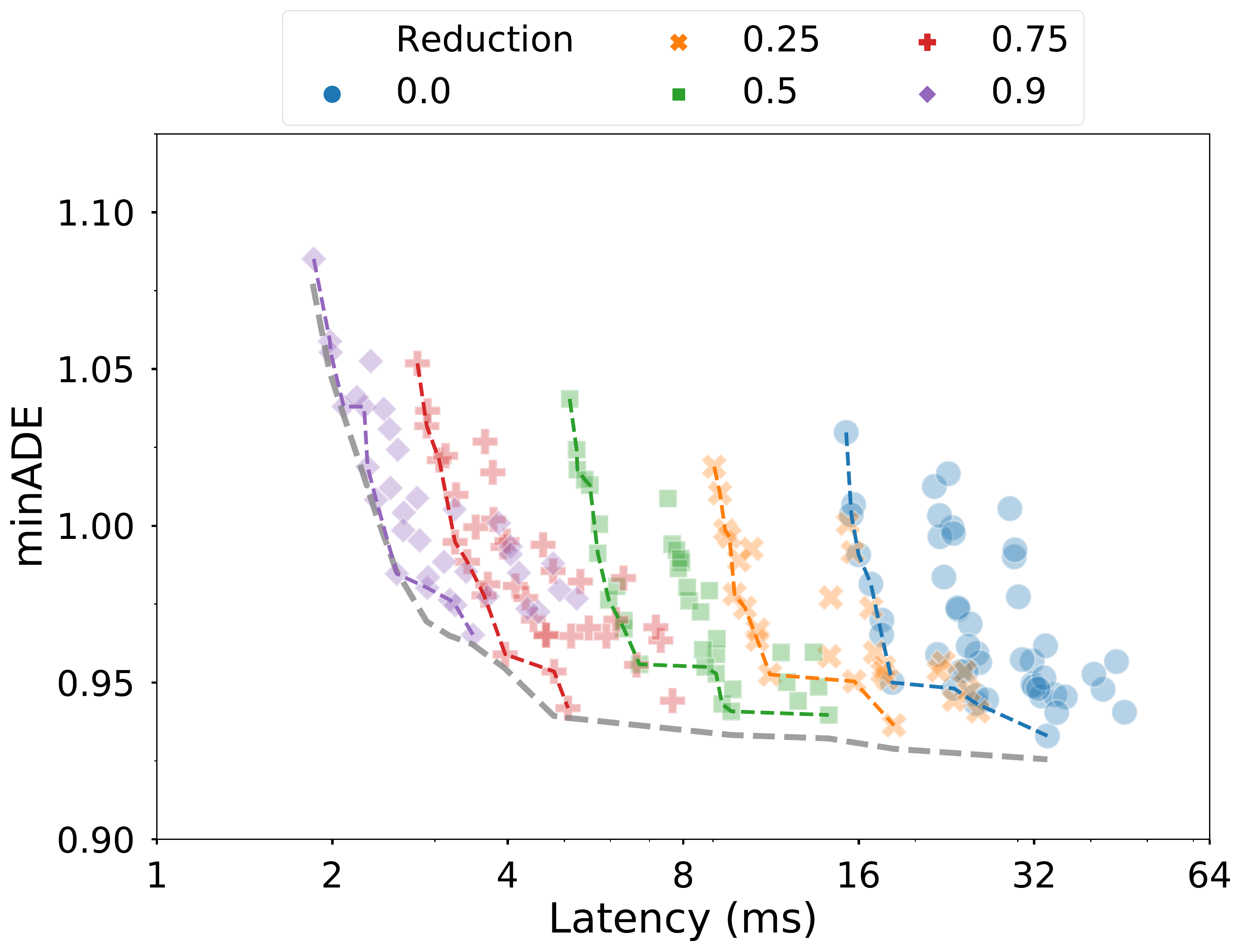}
    \caption{Hierarchical Fusion.}
    \label{fig:hier-lq-results}
  \end{subfigure}

\caption{Latent queries reduce models' latency without significant degradation to the quality.}
  \label{fig:lq-results}
\end{figure}

Figure \ref{fig:lq-results} shows the results of applying latent queries, which speeds up all fusion models by 2x-16x times with minimal to no quality regression.
Early and hierarchical fusion still produce the best quality results, showing the importance of the cross modal interaction stage.

\subsection{Benchmark Results}
\label{sec:sota}
We validate our learnings by comparing Wayformer models to competitive models on popular benchmarks of motion forecasting.
We choose early fusion models since they match the quality of the hierarchical models without increased complexity of implementation.
Moreover, as models' capacity increases they are less sensitive to the choice of fusion (See Figure \ref{fig:fusion-result-params}).
We use latent queries since they speed up models without noticeable quality regression and, in some models, we combine them with factorized attention (see Appendix~\ref{app:lq-factorized_attn}) since that improves the quality further.
We further apply ensembling, a standard practice for producing SOTA results for leaderboard submissions.
Full hyperparameters for \name{} models reported on benchmarks are reported in Appendix \ref{sec:sota-hyperparams}.

When ensembling for WOMD, the model has a single shared encoder but uses $N=3$ separate Transformer decoders.
To merge predictions over the ensemble, we simply combine all mixture components from each predictor to get a total of $N \times 64$ modes, and renormalize the mixture probabilities. 
We then apply our trajectory aggregation scheme (section \ref{subsec:nms}) to the combined mixture distribution to reduce the number of output modes to the desired count $k=6$.

In Table \ref{tab:sota}, we present results on the Waymo Open Motion Dataset and Argoverse Dataset. We use the standard metrics used for the each dataset for their respective evaluation (see Appendix~\ref{app:metrics}).
For the Waymo Open Motion Dataset, both Wayformer early fusion models outperform other models across all metrics; early fusion of input modalities results in better overall metrics independent of the attention structure (multi-axis or factorized attention). 

For Argoverse leaderboard, we train 15 replicas  each with its own encoder and $N=10$ transformer decoders.
To merge predictions over $N$ decoders we follow the aggregation scheme in section \ref{subsec:nms} to result in $k = 6$ modes for each model. We then ensemble 15 such replicas following the same aggregation scheme (section \ref{subsec:nms}) to reduce $N \times 6$ modes to $k = 6$. 

\hide{
\begin{table}[tbh]
\centering
\begin{subtable}[t]{0.49\textwidth}
\vfill
\resizebox{\textwidth}{!}{
\begin{tabular}{ll|lllll} 
\toprule
\multicolumn{7}{c}{\textbf{Waymo Open Motion Dataset}}  \\ 
\midrule
\multicolumn{2}{l|}{Models} & minFDE ($\downarrow$) & minADE  ($\downarrow$) & MR  ($\downarrow$)  & Overlap ($\downarrow$) & mAP$^{*}$ ($\uparrow$)  \\ 
\midrule
\multicolumn{2}{l|}{SimpleCNNOnRaster \cite{konev2021motioncnn}} & 
1.494 &
0.740 &
0.209 &
0.156 &
0.214 \\
\multicolumn{2}{l|}{SceneTransformer \cite{SceneTA}} & 1.212 & 0.612 & 0.156 & 0.147 & 0.279  \\
\multicolumn{2}{l|}{HDGT \cite{hdgt}} & 1.206 & 0.593 &  0.151 & - & 0.285 \\
\multicolumn{2}{l|}{AIR \cite{air2}} & 0.829 & 0.502 & 0.185 & 0.033 & 0.315 \\
\multicolumn{2}{l|}{DenseTNT \cite{gu21dense_tnt}} & 1.551 & 1.039 & 0.157 & 0.178   & 0.328 \\
\multicolumn{2}{l|}{ReCoAt \cite{recoat}} & 0.773  & 0.403  & 0.166  & 0.032 & 0.348  \\
\multicolumn{2}{l|}{Kraken-NMS (Yandex SDG)} & 1.395 & 0.673 & 0.185 & 0.142 & 0.365 \\
\multicolumn{2}{l|}{MultiPath \cite{multipath}} & 2.040 & 0.880 & 0.345 & 0.166 & 0.409 \\
\multicolumn{2}{l|}{MultiPath++ \cite{multipathpp}} & 1.158  & 0.556  & 0.134  & 0.131 & 0.409  \\

\midrule
\midrule
\multicolumn{1}{l}{\multirow{3}{*}{\begin{tabular}{c}Wayformer\\ Early Fusion\end{tabular}}} & \textbf{Attention}  &  & &  &  &  \\ 
\cline{2-2}
 & LQ $+$ Multi-Axis & 1.128 & \textbf{0.545} & \textbf{0.123} & \textbf{0.127} & \textbf{0.419} \\
& LQ $+$ Factorized & \textbf{1.126} & \textbf{0.545}  & \textbf{0.123} & \textbf{0.127} & 0.412  \\
\bottomrule
\end{tabular}
}

\caption{WOMD 2021 public leaderboard results on the official test set.}
\label{tab:WOMD}
 \end{subtable}%
\begin{subtable}[t]{0.485\textwidth}
\vfill
\resizebox{\textwidth}{!}{
\begin{tabular}{ll|llll}
\toprule
\multicolumn{6}{c}{\textbf{Argoverse Leaderboard} ($k=6, d=2m, t=3s$)}                             \\ 
\multicolumn{2}{l|}{Models}                      & Brier-minFDE$^{*}$ ($\downarrow$) & minFDE ($\downarrow$) & MR ($\downarrow$)     & minADE ($\downarrow$)  \\ 
\midrule
\multicolumn{2}{l|}{LaneConv \cite{LaneConv}}                    & 2.0539       & 1.3622 & 0.1600 & 0.8703  \\
\multicolumn{2}{l|}{LaneRCNN \cite{LaneRCNN}}                    & 2.1470       & 1.4526 & 0.1232 & 0.9038  \\
\multicolumn{2}{l|}{mmTransformer \cite{StackedTransformer}}               & 2.0328       & 1.3383 & 0.1540 & 0.8346  \\
\multicolumn{2}{l|}{SceneTransformer \cite{SceneTA}}            & 1.8868       & 1.2321 & 0.1255 & 0.8026  \\
\multicolumn{2}{l|}{TNT \cite{zhao2020tnt}}                         & 2.1401       & 1.4457 & 0.1300 & 0.9400  \\
\multicolumn{2}{l|}{DenseTNT \cite{gu21dense_tnt}}                    & 1.9759       & 1.2858 & 0.1285 & 0.8817  \\
\multicolumn{2}{l|}{TPCN \cite{ye2021tpcn}}                        & 1.9286       & 1.2422 & 0.1333 & 0.8153  \\
\multicolumn{2}{l|}{MultiPath++ \cite{multipathpp}}                 & 1.7932       & 1.2144 & 0.1324 & 0.7897  \\
\multicolumn{2}{l|}{DCMS \cite{ye2022dcms}}                        & 1.7564       & \textbf{1.1350} & \textbf{0.1094} & \textbf{0.7659}  \\ 
\midrule
\midrule
\multirow{3}{*}{
\begin{tabular}{c}Wayformer\\ Early Fusion\end{tabular}
}  & \textbf{Attention}  &               &        &        &         \\ 
\cline{2-2}
                            & LQ + Multi-Axis &    \textbf{1.7408}          & 1.1615       &   0.1186     &  0.7675       \\
                            & LQ + Factorized & 1.7451       & 1.1625 & 0.1192 & 0.7672  \\
\bottomrule
\end{tabular}
}
\caption{Argoverse 2021 leaderboard results. The leaderboard ranks methods using Brier-minFDE, in which both Wayformer variants outperform all prior methods. \todo{NN: include Polkach?}, \todo{NN: add appendix for LQ + factorized}}
    \label{tab:argoverse}
\end{subtable}

\caption{Results of Wayformer - Early Fusion models with select state-of-the-art baselines.
}
\end{table}
}

\begin{table}[tbh]
    \centering
    \resizebox{\textwidth}{!}{
    \begin{tabular}{ll|lllll|llll}
\toprule
\multicolumn{2}{l}{} & \multicolumn{5}{|c|}{\textbf{Waymo Open Motion Dataset}} & \multicolumn{4}{c}{\textbf{Argoverse Dataset}}  \\ 
\multicolumn{2}{l|}{Models} & minFDE ($\downarrow$) & minADE  ($\downarrow$) & MR  ($\downarrow$)  & Overlap ($\downarrow$) & mAP$^{*}$ ($\uparrow$) &
Brier-minFDE$^{*}$ ($\downarrow$) & minFDE ($\downarrow$) & MR ($\downarrow$)     & minADE ($\downarrow$)  \\ 
\midrule
\multicolumn{2}{l|}{SceneTransformer \cite{SceneTA}} & 1.212 & 0.612 & 0.156 & 0.147 & 0.279 & 1.8868       & 1.2321 & 0.1255 & 0.8026  \\
\multicolumn{2}{l|}{DenseTNT \cite{gu21dense_tnt}} & 1.551 & 1.039 & 0.157 & 0.178   & 0.328 & 1.9759       & 1.2858 & 0.1285 & 0.8817 \\
\multicolumn{2}{l|}{MultiPath \cite{multipath}} & 2.040 & 0.880 & 0.345 & 0.166 & 0.409 & - & - & - & - \\
\multicolumn{2}{l|}{MultiPath++ \cite{multipathpp}} & 1.158  & 0.556  & 0.134  & 0.131 & 0.409 & 1.7932       & 1.2144 & 0.1324 & 0.7897  \\
\multicolumn{2}{l|}{LaneConv} & - & - & - & - & -  &                      2.0539       & 1.3622 & 0.1600 & 0.8703  \\
\multicolumn{2}{l|}{LaneRCNN \cite{LaneRCNN}} & - & - & - & - & - &                     2.1470       & 1.4526 & 0.1232 & 0.9038  \\
\multicolumn{2}{l|}{mmTransformer \cite{StackedTransformer}} & - & - & - & - & -               & 2.0328       & 1.3383 & 0.1540 & 0.8346 \\
\multicolumn{2}{l|}{TNT \cite{zhao2020tnt}}  & - & - & - & - & -                        & 2.1401       & 1.4457 & 0.1300 & 0.9400 \\
\multicolumn{2}{l|}{DCMS \cite{ye2022dcms}}      & - & - & - & - & -                   & 1.7564       & \textbf{1.1350} & \textbf{0.1094} & \textbf{0.7659}  \\ 
\midrule
\multicolumn{1}{l}{\multirow{3}{*}{\begin{tabular}{c}Wayformer\\ Early Fusion\end{tabular}}} & \textbf{Attention}  &  &  &  &  & & & & & \\ 
\cline{2-2}
 & LQ $+$ Multi-Axis & 1.128 & \textbf{0.545} & \textbf{0.123} & \textbf{0.127} & \textbf{0.419} & \textbf{1.7408}          & 1.1615       &   0.1186     &  0.7675 \\
& LQ $+$ Factorized & \textbf{1.126} & \textbf{0.545}  & \textbf{0.123} & \textbf{0.127} & 0.412 & 1.7451       & 1.1625 & 0.1192 & 0.7672 \\
\bottomrule
\end{tabular}

    }
    \caption{Wayformer models and select SOTA baselines on Waymo Open Motion Dataset 2021 and Argoverse 2021. \textbf{*} denotes the metric used for leaderboard ranking. LQ denotes latent query.}
    \label{tab:sota}
\end{table}

\section{Related Work}

\paragraph{Motion prediction architectures :}

Increasing interest in self-driving applications and the availability of benchmarks \cite{argoverse,lyft5,WOMD} has allowed motion prediction models to flourish.
Successful modeling techniques fuse multi-modal inputs that represent different static, dynamic, social and temporal aspects of the scene. 
One class of models draws heavily from the computer vision literature, rendering inputs as a multichannel rasterized top-down image \cite{cui2019mtp,multipath,DBLP:journals/corr/LeeCVCTC17,DBLP:journals/corr/abs-1906-08945,casas2018intentnet,zhao2020tnt}.
In this approach, relationships between scene elements are rendered in the top down orthographic plane and modeled via spatio-temporal convolutional networks.
However, the localized structure of convolutions is well suited to processing image inputs, but is not effective at capturing the long range spatio-temporal relationships.
A popular alternative is to use an entity-centric approach, where agent state history is typically encoded via sequence modeling techniques like RNNs \cite{multiheadAF,wimp2020,social-lstm,rhinehart2019precog} or temporal convolutions \cite{liang2020laneGCN}.
Road elements are approximated with basic primitives (e.g. piece-wise linear segments) which encode pose and semantic information.
Modeling relationships between entities is often presented as an information aggregation process, and models employ pooling \cite{zhao2020tnt,gao2020vectornet,social-lstm,social-gan,multiheadAF,DBLP:journals/corr/LeeCVCTC17}, soft-attention \cite{multiheadAF,zhao2020tnt} or graph neural networks \cite{casas2020spagnn,liang2020laneGCN,wimp2020}.
Like our proposed method, several recent models use Transformers \cite{vaswani2017attention}, which are a popular state-of-the-art choice for sequence modeling in NLP \cite{devlin2018bert,DBLP:journals/corr/abs-2005-14165}, and have  shown promise in core computer vision tasks such as detection \cite{carion20detr,DBLP:journals/corr/abs-1904-09925,DBLP:journals/corr/abs-2101-11605}, tracking \cite{DBLP:journals/corr/abs-2008-07725} and classification \cite{DBLP:journals/corr/abs-1904-09925,DBLP:journals/corr/abs-1906-05909}.

\paragraph{Iterative cross-attention}
A recent approach to encode multi-modal data is to sequentially process one modality at a time \cite{StackedTransformer,perceiver,Autobots}.
\cite{StackedTransformer} ingests the scene in the order \{agent history, nearby agents, map\}; they argue that it is computationally expensive to perform self-attention over multiple modalities at once.
\cite{Autobots} pre-encodes the agent history and contextual agents through self-attention and cross-attends to the map with agent encodings as queries.
The order of self-attention and cross-attention relies heavily on the designer's intuition and has, to our knowledge, not been ablated before.

\paragraph{Factorized Attention}
Flattening high dimensional data leads to long sequences which make  self-attention computationally prohibitive.
\cite{axialattn} proposed limiting each attention operation to a single axes to alleviate the computational costs and applied this technique to autoregressive generative modeling for images.
Similarly, \cite{vivit} factorize the spatial and temporal dimensions of the video input when constructing their self-attention based classifier.
This axis based attention, which gets applied in interleaved fashion across layers, has been adopted in Transformer-based motion forecasting models \cite{Autobots} and graph neural network approaches \cite{SpatioTemporalGT}.
The order of applying attention over \{temporal, social/spatial\} dimensions has been studied with two different common patterns: (a) Temporal first \cite{social-lstm,social-gan,bigat} (b) Social/Spatial first \cite{stgat,trajectronpp}.
In Section \ref{sec:partial}, we study a `sequential' mode 
and contrast it with interleaved mode where interleave dimensions of attention similar to \cite{Autobots}. 

\paragraph{Multimodal Encoding}
\cite{agentformer} argued that attending to temporal and spatial dimensions independently leads to loss of information.
Moreover, allowing all inputs to self-attend to each other early on the encoding process reduces complexity and the need to handcraft architectures to address the scaling of computation for transformers with the increase in the input sequence length \cite{perceiver-io}.
However, self-attention is known to be computationally expensive for large inputs \cite{transformerbenchmark}, and recently there has been huge interest in approaches improving its scalability.
For a complete discussion of previous works, we refer the reader to the comprehensive survey \cite{efficientTransformers}.
One compelling approach is to use learned latent queries to decouples the number of query vectors of a Transformer encoder from the original input sequence length \cite{settransformer}.
This allows us to set the resolution of the Transformer output to arbitrary scales independent of the input, and flexibly tune model computational costs.
This approach is appealing since it does not assume any structure in the input and has proven effective in fusing multimodal inputs \cite{perceiver-io}. We take inspiration from such frameworks and present a study of their benefits when applied to the task of motion forecasting in the self-driving domain.

\section{Limitations}
\label{sec:limitations}
Scope of the current study is subject to the following limitations:
    (1) Ego-centric modeling is subject to repeated computations on dense scenes.
    This can be alleviated by encoding the scene only once in a global frame of reference.
    (2) Our system input is a sparse abstract state description of the world, which fails to capture some important nuances in highly interactive scenes, e.g., visual cues from pedestrians or fine-granularity contour or wheel angle information for vehicles.
    Learning perception and prediction end-to-end could unlock improvements.
    (3) We model the distribution over possible futures independently per agent, and temporally conditionally independent for each agent given intent.
    These simplifying assumptions allow for efficient computation but fail to fully describe combinatorially many futures.
    Multi-agent, temporally causal models could show further benefits in interactive situations.

\clearpage
\acknowledgments{We thank Balakrishnan Varadarajan for help on ensembling strategies; Dragomir Anguelov and Eugene Ie for their helpful feedback on the paper.}


\bibliography{example}  

\newpage
\label{sec:appendix}
\appendix
\section*{Appendix}
\section{Factorized Latent Query Attention}
\label{app:lq-factorized_attn}

\begin{figure}[h]
  \begin{subfigure}[b]{0.71\columnwidth}
    \includegraphics[width=\linewidth]{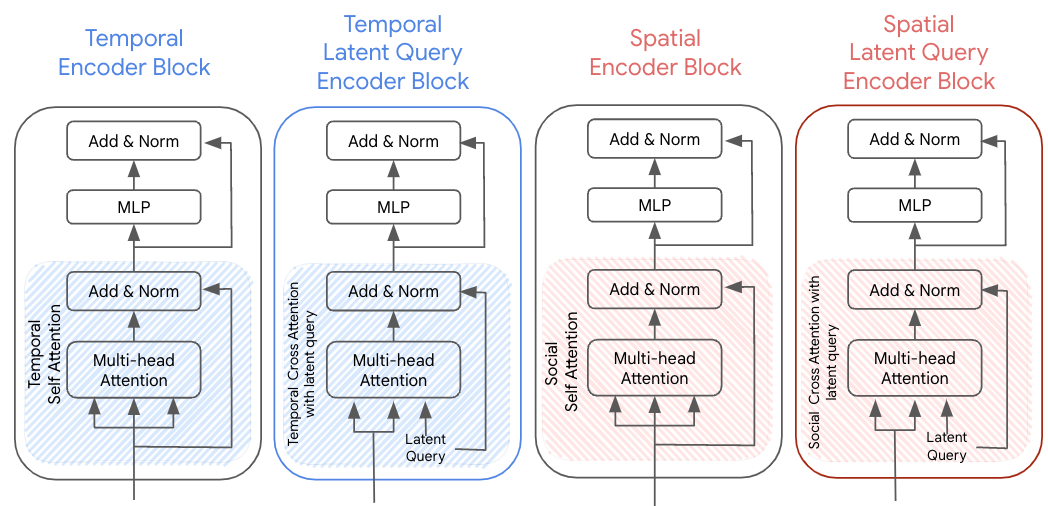}
    \caption{Encoder Blocks}
    \label{fig:lq-factorized-attn-encoder-blocks}
  \end{subfigure}
  \begin{subfigure}[b]{0.28\columnwidth}
    \includegraphics[width=\linewidth]{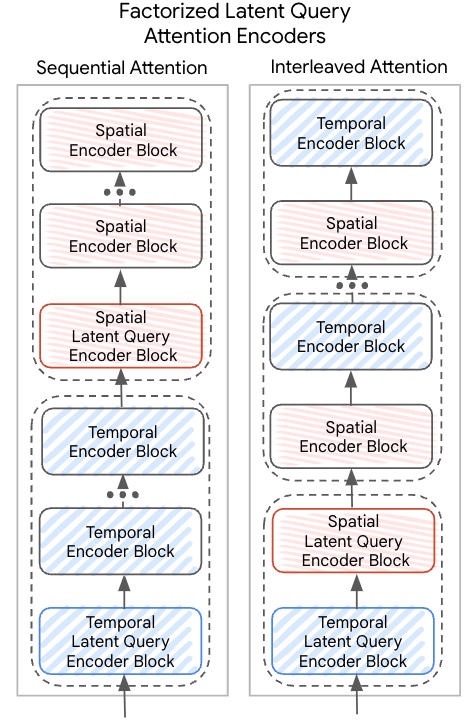}
    \caption{Encoders}
    \label{fig:lq-factorized-attn-encoders}
  \end{subfigure}
  \caption{A summary of encoder architectures considered for Wayformer. (a) provides an overview of different encoder blocks and (b) explains how these blocks are arranged to construct the encoder.}
  \label{fig:lq-factorized-attn}
\end{figure}

Figure~\ref{fig:lq-factorized-attn-encoder-blocks} shows the implementation of Factorized latent query attention encoder blocks and Figure~\ref{fig:lq-factorized-attn-encoders} shows how they are used in constructing the encoders. Specifically in factorized attention (sequential or interleaved), the first temporal encoder block and the first spatial encoder blocks in Figure~\ref{fig:blocks-encoders} are replaced with temporal latent query encoder block and spatial latent query encoder block respectively.

\section{Hyperparameters}
\label{app:hyper}
\begin{table}[h]
    \centering
    \begin{tabular}{l|l}
        \toprule
        \textbf{Hyperparameter} & \textbf{Values} \\
        \midrule
        Hidden size & \{128, 256, 512\} \\
        Intermediate size & \{2x, 4x\} \text{ hidden size} \\
        Num encoder layers & [2, 16] \\
        Num decoder layers & [2, 16] \\
        Latent query ratio & \{0.25, 0.5, 0.75 1.0\} \\
        Number GMM modes & 64 \\
        Optimizer & AdamW \\
        Initial learning rate & 2e-4 \\
        Training steps & 1000000 \\
        Learning rate decay & linear \\
        Batch size & 256 \\
        \bottomrule
    \end{tabular}
    \caption{Model and training hyperparameters across all ablation experiments done on WOMD. }
    \label{tab:hyperparams}
\end{table}

\vspace{3mm}
\begin{table}[h]
    \centering
    \begin{tabular}{l|l|l}
    \toprule
        \textbf{Hyperparameter} & \textbf{WOMD} & \textbf{Argoverse} \\
        \midrule
        Max num history timesteps  & 11  & 20 \\
        (including current timestep) & & \\
        Max num roadgraph feats & 512 
        & 1024\\
        Max num context agents & 64 & 64\\
        Max num traffic lights & 32 & 32\\
        \bottomrule
    \end{tabular}
    \caption{Hyperparameters for generating WOMD and Arogverse input features. Fixed for all experiments}
    \label{tab:hyperparams_features}
\end{table}

\section{Trajectory Aggregation Details}
\label{app:nms}

Given a distance threshold $D$, the trajectory aggregation scheme attempts to first select the fewest centroid modes such that all output modes are within a final distance $D$ away from the nearest centroid. 
The aggregation algorithm iteratively selects centroid modes by greedily selecting the output mode that covers the maximum total likelihood out of the uncovered modes, and proceeds until all output modes have been covered.

After initializing these $k$ centroid modes, the aggregation algorithm then proceeds into a refinement stage and runs another iterative procedure similar to $k$-means clustering starting from the initial centroid modes. In each iteration, each centroid mode becomes of the weighted average of all output modes assigned to it, and then output modes are reassigned to the new closest centroid mode.

\section{SOTA Wayformer Details}
We describe the the hyperparameters used for WOMD and Argoverse benchmark results in Tables \ref{tab:womd-sota-hyperparams} and \ref{tab:argo-sota-hyperparams} respectively.
\label{sec:sota-hyperparams}
\begin{table}[h]
    \centering
    \begin{tabular}{l|l|l}
        \toprule
        \textbf{Hyperparameter} & \textbf{Multi-axis Latent Query} & \textbf{Factorized Latent Query} \\
        \midrule
        Hidden size & 256 & 256\\
        Intermediate size & 1024 & 1024  \\
        Num encoder layers & 2 & 4\\
        Num decoder layers & 8 & 4 \\
        Latent queries & 192 & 4 time latents, 192 spatial latents \\
        Number GMM modes & 64 & 64 \\
        Ensemble size & 3 & 3 \\
        Optimizer & AdamW & AdamW \\
        Initial learning rate & 2e-4 & 2e-4 \\
        Learning rate decay & linear & linear\\
        Training steps & 1200000 & 1000000 \\
        Batch size & 256 & 256\\
        Aggregation initial distance threshold & 2.3 & 2.3\\
        Aggregation refinement iterations & 3 &  3 \\
        Aggregation max num trajectories & 6 & 6 \\
        \bottomrule
    \end{tabular}
    \caption{Model and training hyperparameters for benchmark experiments on Waymo Open Motion 2021 Dataset}
    \label{tab:womd-sota-hyperparams}
\end{table}

\begin{table}[h]
    \centering
    \begin{tabular}{l|l|l}
        \toprule
        \textbf{Hyperparameter} & \textbf{Multi-axis Latent Query} & \textbf{Factorized Latent Query} \\
        \midrule
        Encoder Hidden size & 128 & 256\\
        Encoder Intermediate size & 512 & 1536  \\
        Decoder Hidden size & 128 & 128\\
        Decoder Intermediate size & 512 & 768  \\
        Num encoder layers & 4 & 4\\
        Num decoder layers & 6 & 6 \\
        Latent queries & 1024 & 6 time latents, 192 spatial latents \\
        Number GMM modes & 6 & 6 \\
        Ensemble size & 10 & 10 \\
        Optimizer & AdamW & AdamW \\
        Initial learning rate & 2e-4 & 2e-4 \\
        Learning rate decay & linear & linear\\
        Training steps & 1000000 & 1000000 \\
        Batch size & 4 & 4\\
        Aggregation initial distance threshold & 2.9 & 2.9\\
        Aggregation refinement iterations & 5 &  5 \\
        Aggregation max num trajectories & 6 & 6 \\
        \bottomrule
    \end{tabular}
    \caption{Model and training hyperparameters for benchmark experiments on Argoverse 2021 Dataset.}
    \label{tab:argo-sota-hyperparams}
\end{table}

\section{Metrics}
\label{app:metrics}
We compare models using competition specific metrics associated with these datasets. For all metrics, we consider only the top $k=6$ most likely modes output by our model (after trajectory aggregation) and use only the mean of each mode.

Specifically, we report the following metrics taken from the evaluation procedure used in the standard evaluations based on the dataset being used.

$\mathbf{minDE_k^t}$ (Minimum Distance Error): Considers the top-$k$ most likely trajectories output by the model, and computes the minimum distance to the ground truth trajectory at timestep $t$.

$\mathbf{MR^t}$ (Miss Rate): For each predicted trajectory, we compute whether it is sufficiently close to the predicted agent's ground truth trajectory at time $t$. Miss rate as the proportion of predicted agents for which none of the predicted trajectories are sufficiently close to the ground truth. We defer details of how a trajectory is determined to be sufficiently close to the WOMD metrics definition \cite{WOMD}.

$\mathbf{minADE_k}$ (Minimum Average Distance Error): Similar to $minDE^t_k$
, but the distance is calculated as an average over
all timesteps.

$\mathbf{mAP}$: For each set of predicted trajectories, we have at most one positive - the one closest to the ground truth and
which is within $\tau$ distance from the ground truth. The other predicted trajectories are reported as misses. From this, we
can compute precision and recall at various thresholds. Following WOMD metrics definition \cite{WOMD} the agents future
trajectories are partitioned into behavior buckets, and an area under the precision-recall curve is computed using the
possible true positive and false positives per agent, giving us Average Precision per behavior bucket. The total mAP
value is a mean over the AP’s for each behavior bucket.

$\mathbf{Overlap^t}$: The fraction of timesteps of the most likely trajectory prediction for which the prediction overlaps with the corresponding timestep real future trajectory of another agent.

$\mathbf{minFDE}$ (Minimum Final Displacement Error): The L2 distance between the endpoint of the best forecasted trajectory and the ground truth.

$\mathbf{brier - minFDE}$: is defined as the sum of minFDE and the brier score $(1 - p)^2$, where $p$
 is the probability of the best-predicted trajectory.

\section{Qualitative Wins}

In this section, we present some examples of Wayformer (WF) predictions on WOMD scenes in comparison with MultiPath++ (MP++) model \cite{multipathpp}.  In all the following examples, (a) Hue indicates time horizon (0s - 8s), while transparency indicates probability. (b) Rectangles indicate vehicles, and squares indicate pedestrians or cyclists.

\begin{figure}[h]
  \begin{subfigure}[b]{0.49\columnwidth}
    \includegraphics[width=\linewidth]{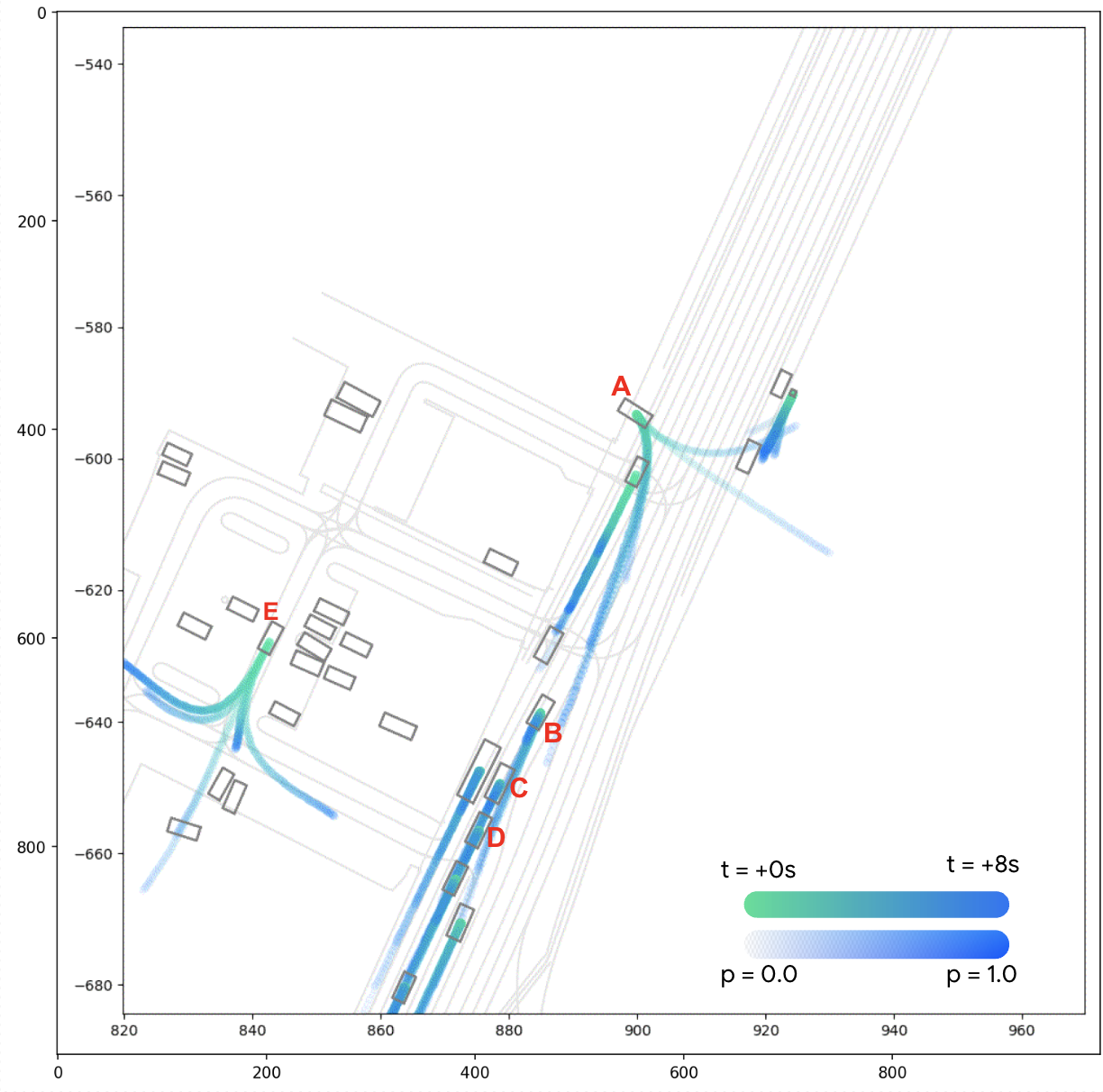}
    \caption{MultiPath++ (MP++)}
  \end{subfigure}
  \begin{subfigure}[b]{0.49\columnwidth}
    \includegraphics[width=\linewidth]{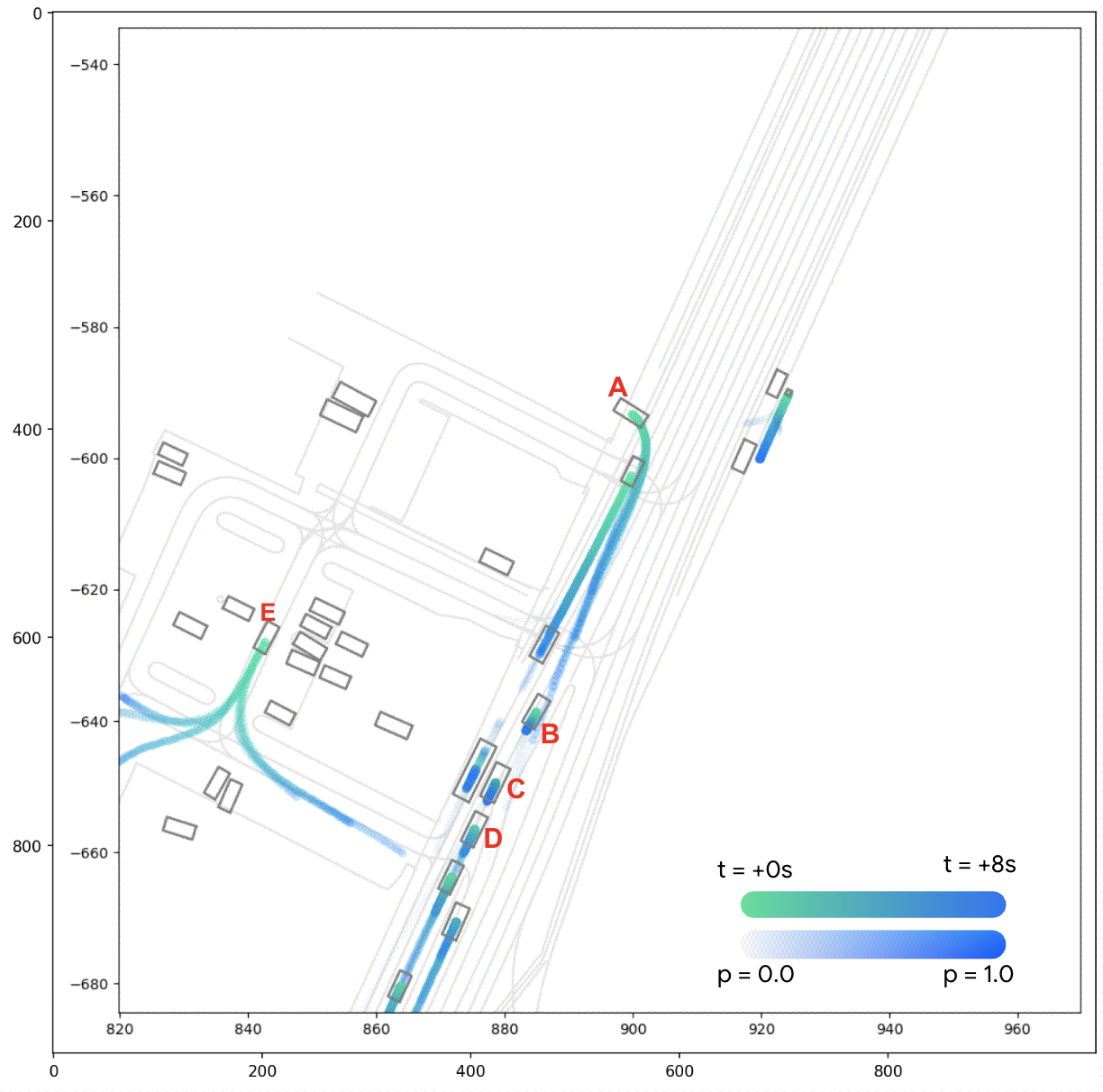}
    \caption{Wayformer (WF)}
  \end{subfigure}
  \caption{This scenario represents a multi-lane road with a parking lot on the left side. Here, we see that WF's performance on several vehicles is more safe and road following than that of MP++. For example: (a) Vehicle A is seen merging onto the road coming out of a parking lot. MP++'s predictions are completely off-road while WF's predictions follow rules of the road. (b) Vehicles B, C, and D's predictions overlap with each other for MP++ predicting collision with each other. But, WF correctly predicts that D yields for the vehicle before, C yields for D and B yields for C. (c) MP++'s predictions for vehicle E navigating the parking lot go through an already parked vehicles, while WF  understands the interactions better and produces predictions which are not colliding.}
  \label{fig:qualitative-wins-0}
\end{figure}

\begin{figure}[h]
  \begin{subfigure}[b]{0.49\columnwidth}
    \includegraphics[width=\linewidth]{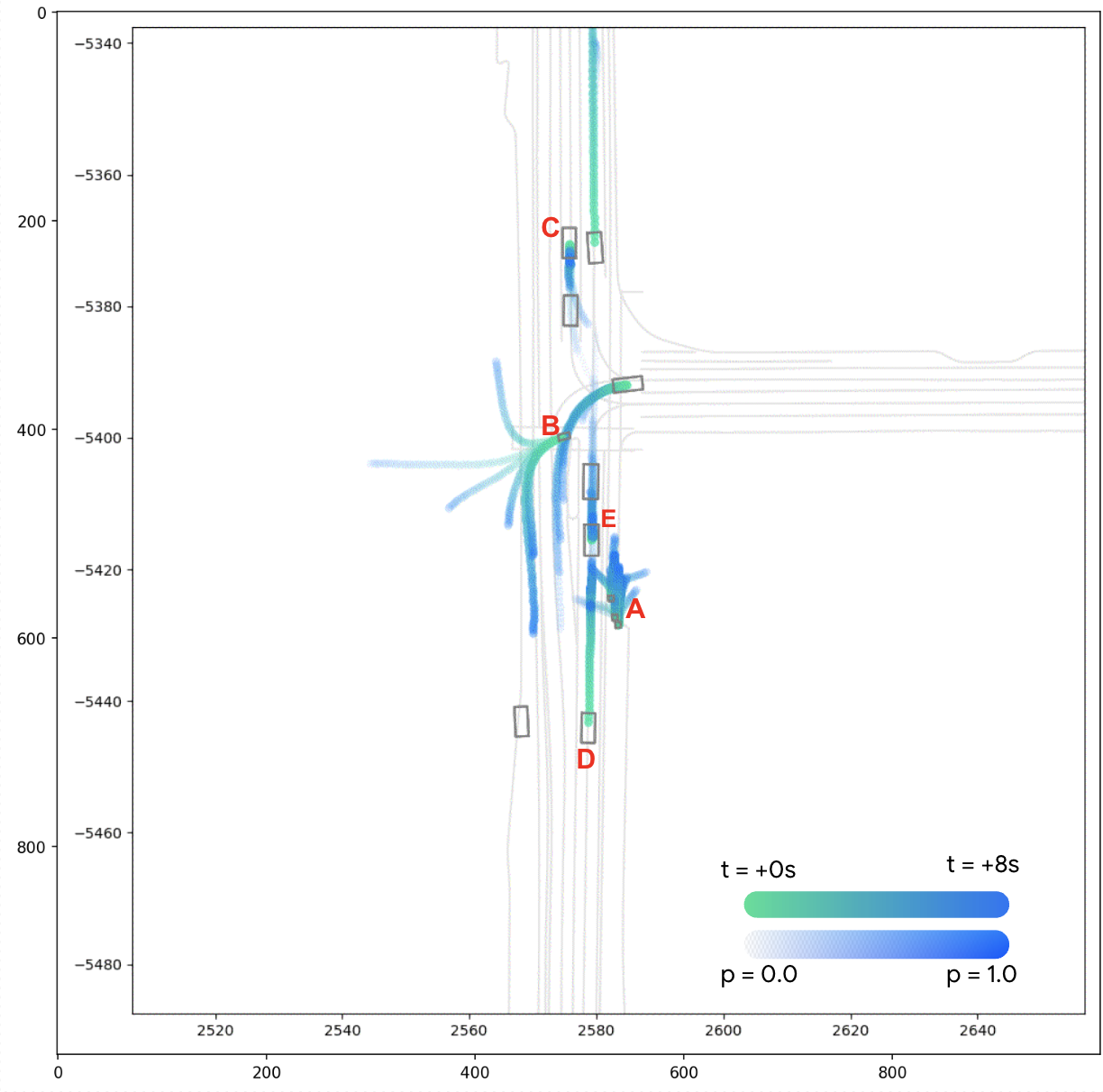}
    \caption{MultiPath++ (MP++)}
  \end{subfigure}
  \begin{subfigure}[b]{0.49\columnwidth}
    \includegraphics[width=\linewidth]{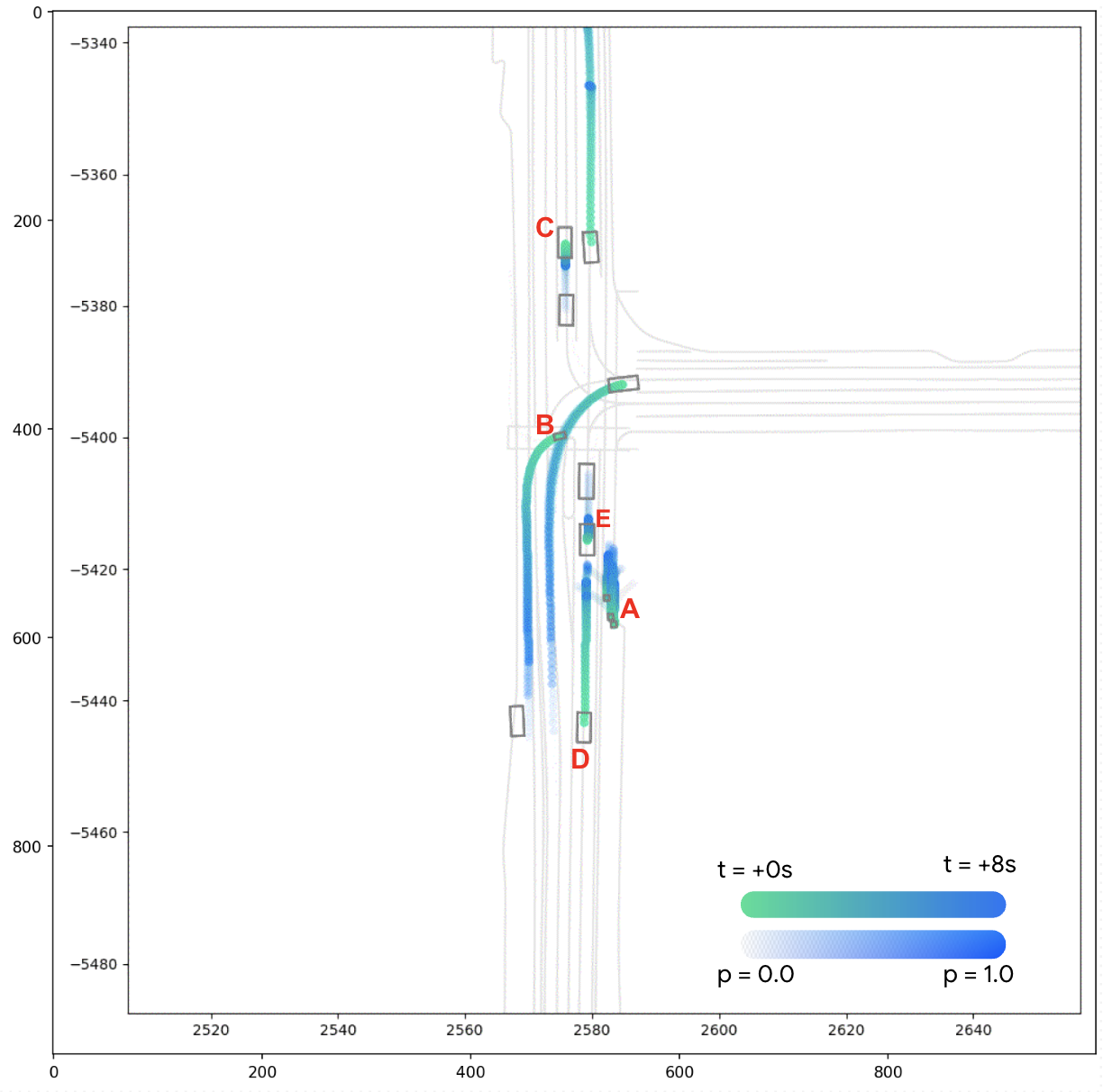}
    \caption{Wayformer (WF)}
  \end{subfigure}
  \caption{This scenario represents a T intersection. Here we see (a) a cyclist B, making a left turn. MP++'s predictions are off-road and going beyond the available road. But, WF's predictions follow rules of the road and present multiple speed profiles for the same action of taking a left turn. (b) We also see better predictions for a pedestrian (pedestrian A) where MP++ predicts that the pedestrian is going to walk onto the road with  oncoming traffic. But, WF's  predictions are  constrained to the side walk. (c) In addition, we also notice that WF's predicts safe futures for vehicles C, D and E in comparison with MP++.}
  \label{fig:qualitative-wins-21}
\end{figure}

\begin{figure}[h]
  \begin{subfigure}[b]{0.49\columnwidth}
    \includegraphics[width=\linewidth]{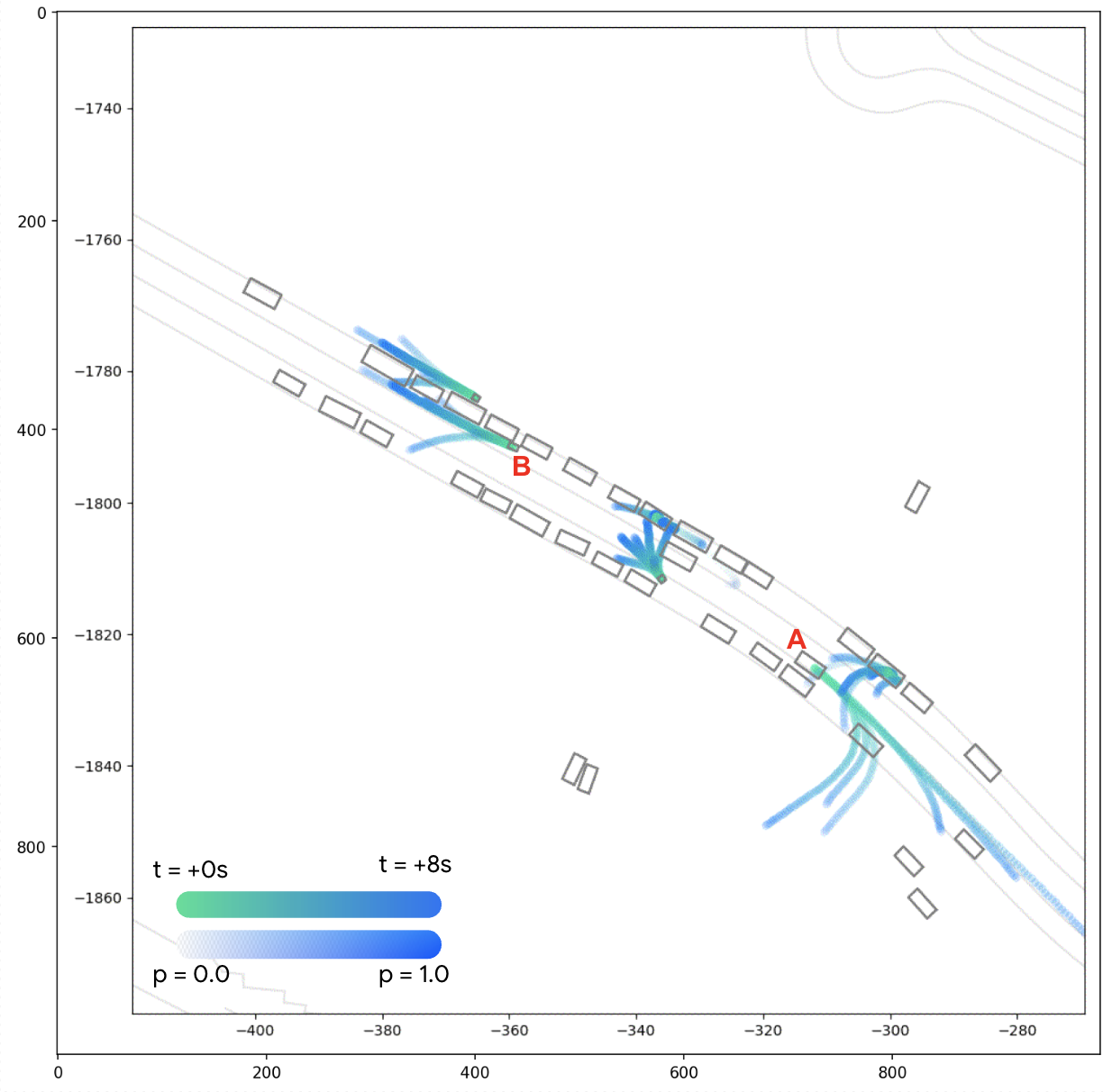}
    \caption{MultiPath++ (MP++)}
  \end{subfigure}
  \begin{subfigure}[b]{0.49\columnwidth}
    \includegraphics[width=\linewidth]{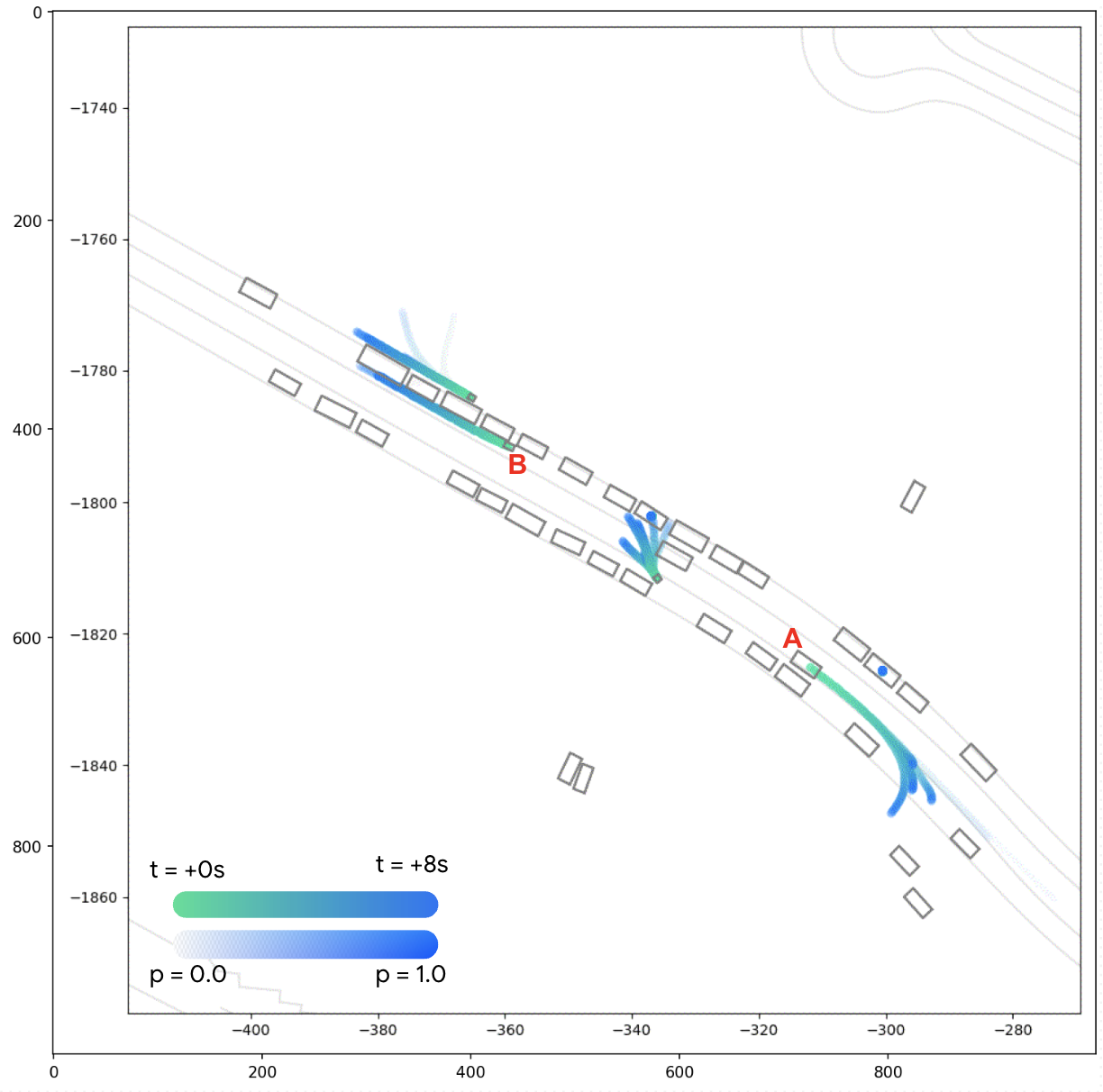}
    \caption{Wayformer (WF)}
  \end{subfigure}
  \caption{This scenario represents a vehicle (agent A) turning into a parking structure. MP++'s prediction discounts the presence of other parked vehicles and some predictions are made through the parked agents. WF models these interactions better and only predicts trajectories that do not collide with other parked entities.}
  \label{fig:qualitative-wins-2}
\end{figure}

\begin{figure}[h]
  \begin{subfigure}[b]{0.49\columnwidth}
    \includegraphics[width=\linewidth]{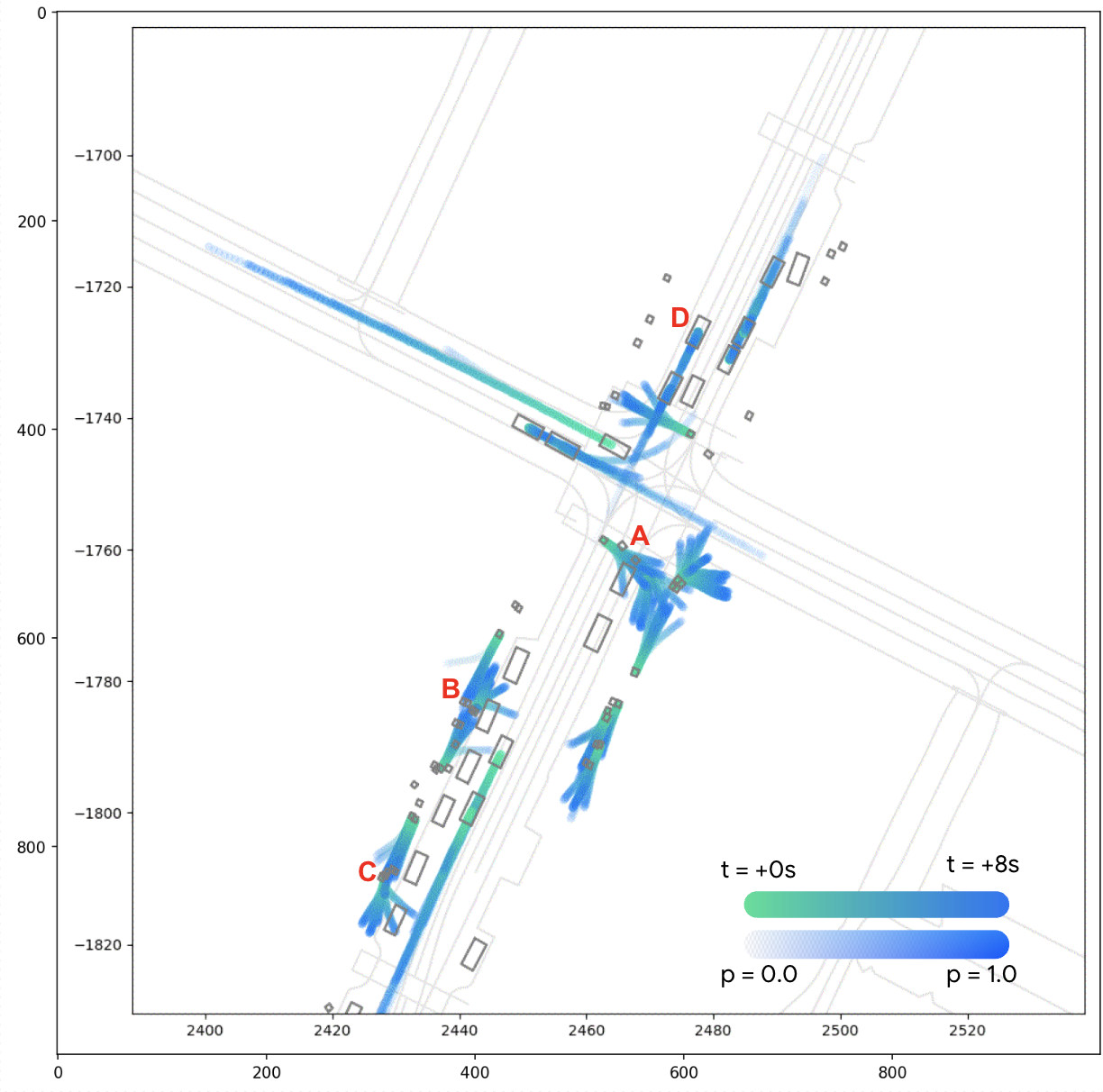}
    \caption{MultiPath++ (MP++)}
  \end{subfigure}
  \begin{subfigure}[b]{0.49\columnwidth}
    \includegraphics[width=\linewidth]{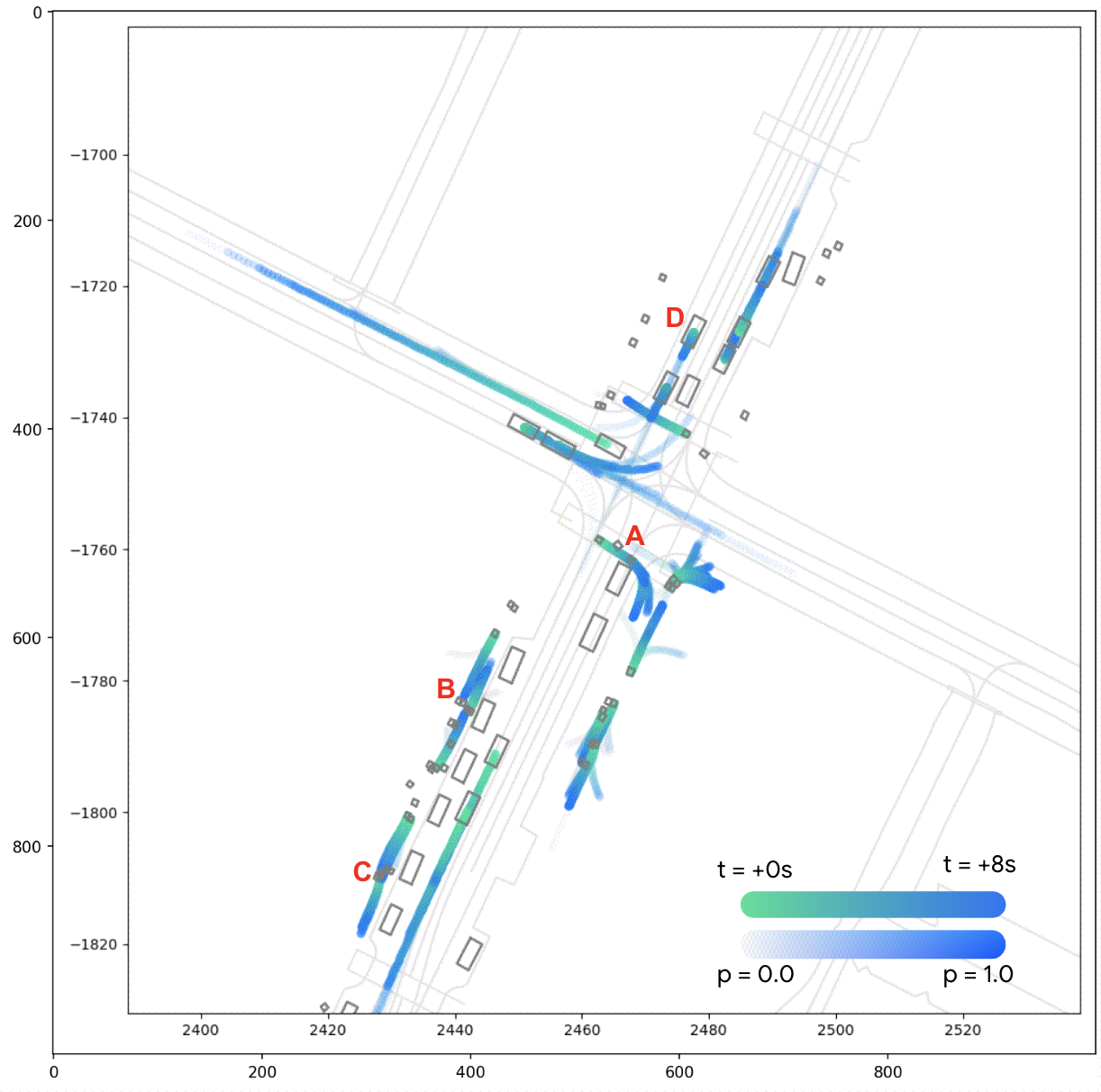}
    \caption{Wayformer (WF)}
  \end{subfigure}
  \caption{This scenario represents a busy 4-way intersection. First we discuss the WF improvements for pedestrian trajectory predictions. MP++ predicts pedestrian (A) as going into the oncoming vehicle demonstrating it fails to model this spatial interaction. WF demonstrates how the same pedestrian crosses in-front of this stopped vehicle and continues to walk on the corsswalk on the opposite side of the road. Pedestrian (agent B and C) on the lower left corner of the image show similar behavior. MP++ predicts them to bump into cars parked right next to them and walk onto the road surface towards oncoming traffic. WF on the other hand predicts nice and consistent along road trajectories for these pedestrians. We now observe the predicts for a vehicle (agent D) in this scene. MP++ predicts the trajectories of this vehicle to collide both with the static car in-front of it as well as the pedestrian passing in-front on that car. WF models all these spatial interactions well and predicts the trajectories for these car to wait behind the car in-front of it and not nudge into the pedestrian crossing in-front. }
  \label{fig:qualitative-wins-3}
\end{figure}

\begin{figure}[h]
  \begin{subfigure}[b]{0.49\columnwidth}
    \includegraphics[width=\linewidth]{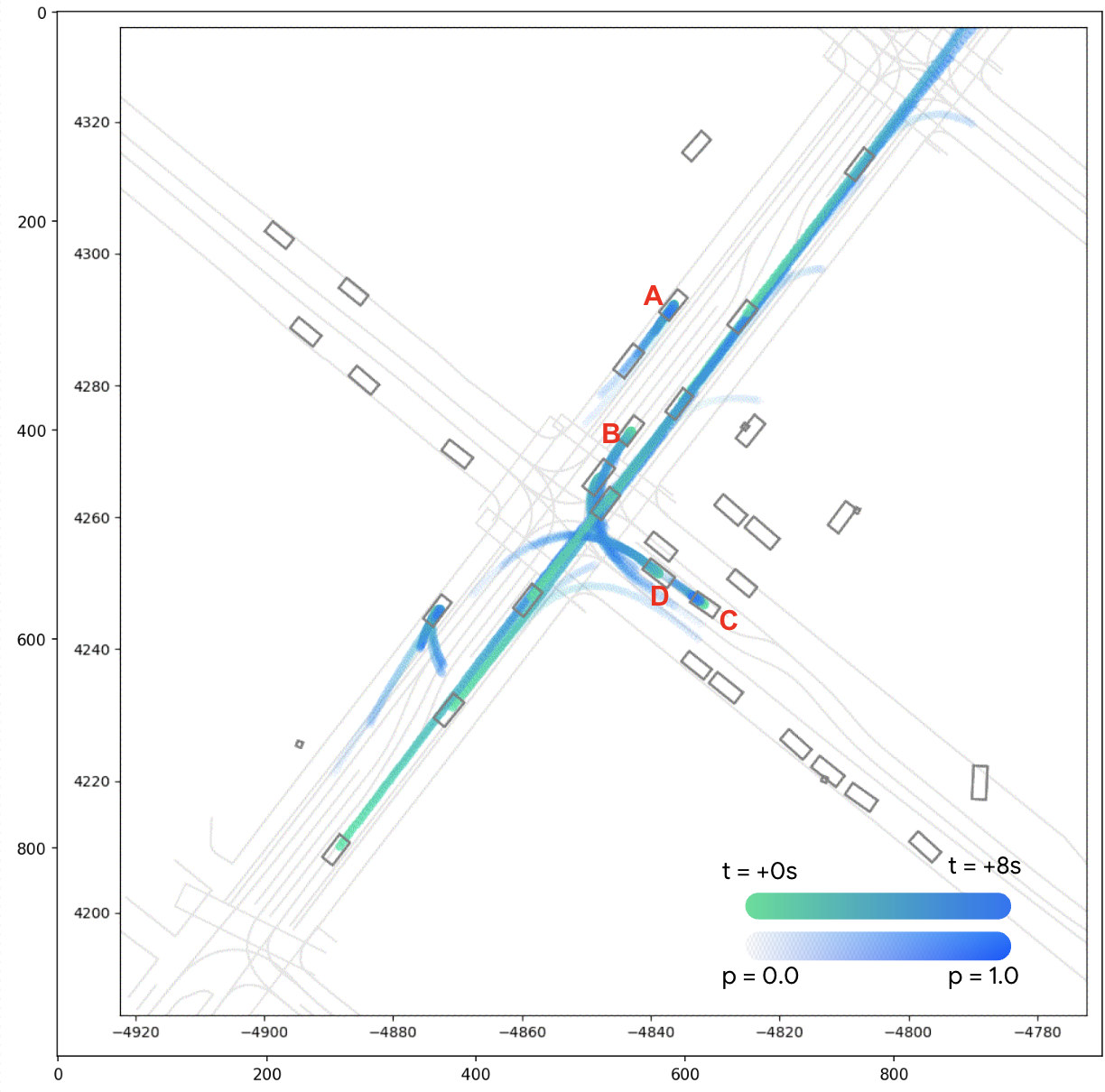}
    \caption{MultiPath++ (MP++)}
  \end{subfigure}
  \begin{subfigure}[b]{0.49\columnwidth}
    \includegraphics[width=\linewidth]{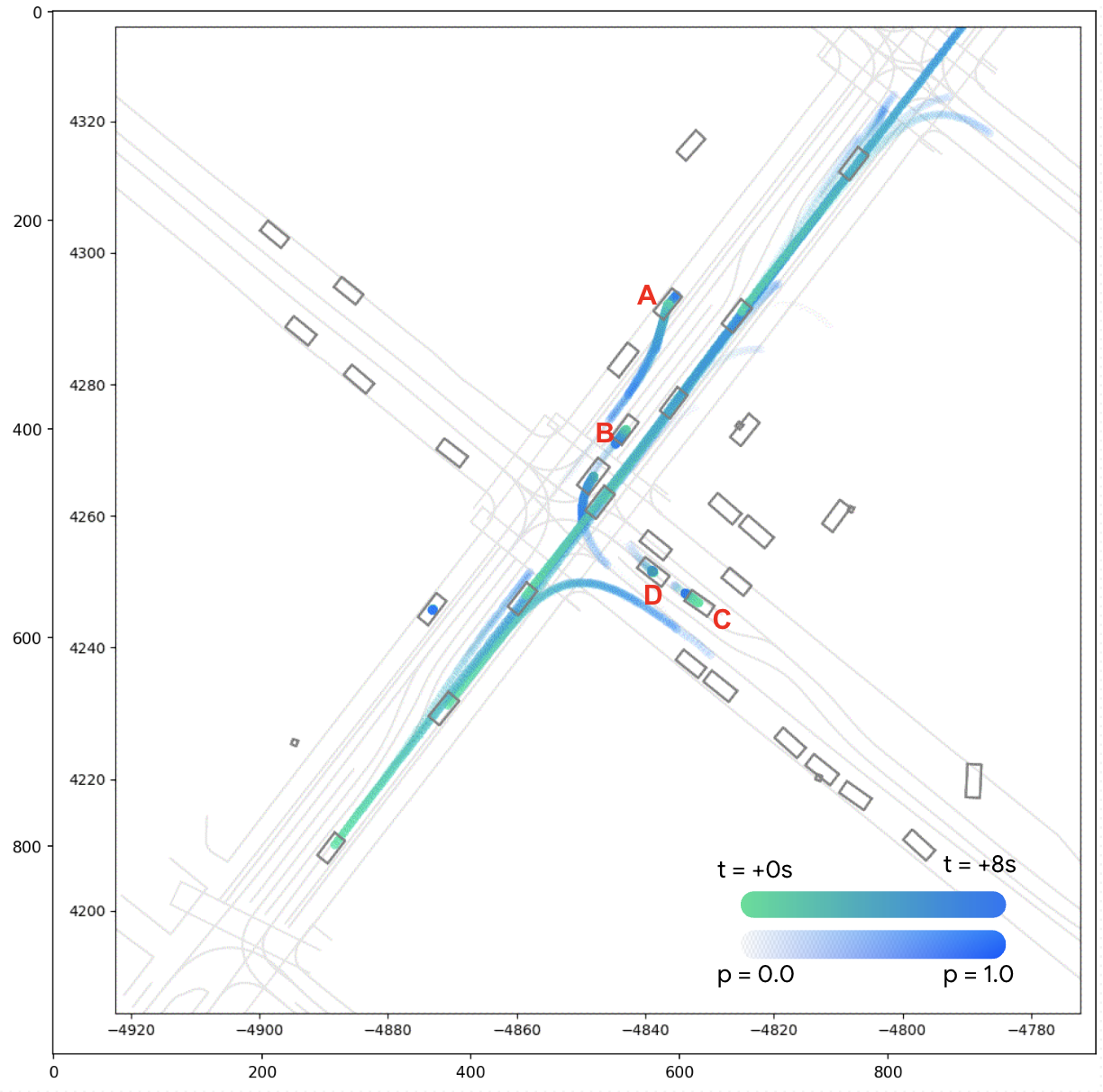}
    \caption{Wayformer (WF)}
  \end{subfigure}
  \caption{This scenario represents a comlpex 4-way intersection with lots of cars passing through. Similar to Fig- \ref{fig:qualitative-wins-3} we see MP++ predicting trajectories for vehicles (agent A, B, C an d D) in the scene to collide with cars in-front of them. WF demonstrates very sophisticated behavior. For agent A, it is able to estimate that the car parked in-front of agent A is a double-parked vehicle and there is space on the road next to it, so it predicts trajectories that nudge around it. For B, C and D it is able to carefully model the rules of the road and allow either oncoming ( in case of agent B)  or cross traffic ( in case of agent C and D) to take precedence and predicts yielding trajectories for them. }
  \label{fig:qualitative-wins-4}
\end{figure}

\begin{figure}[h]
  \begin{subfigure}[b]{0.49\columnwidth}
    \includegraphics[width=\linewidth]{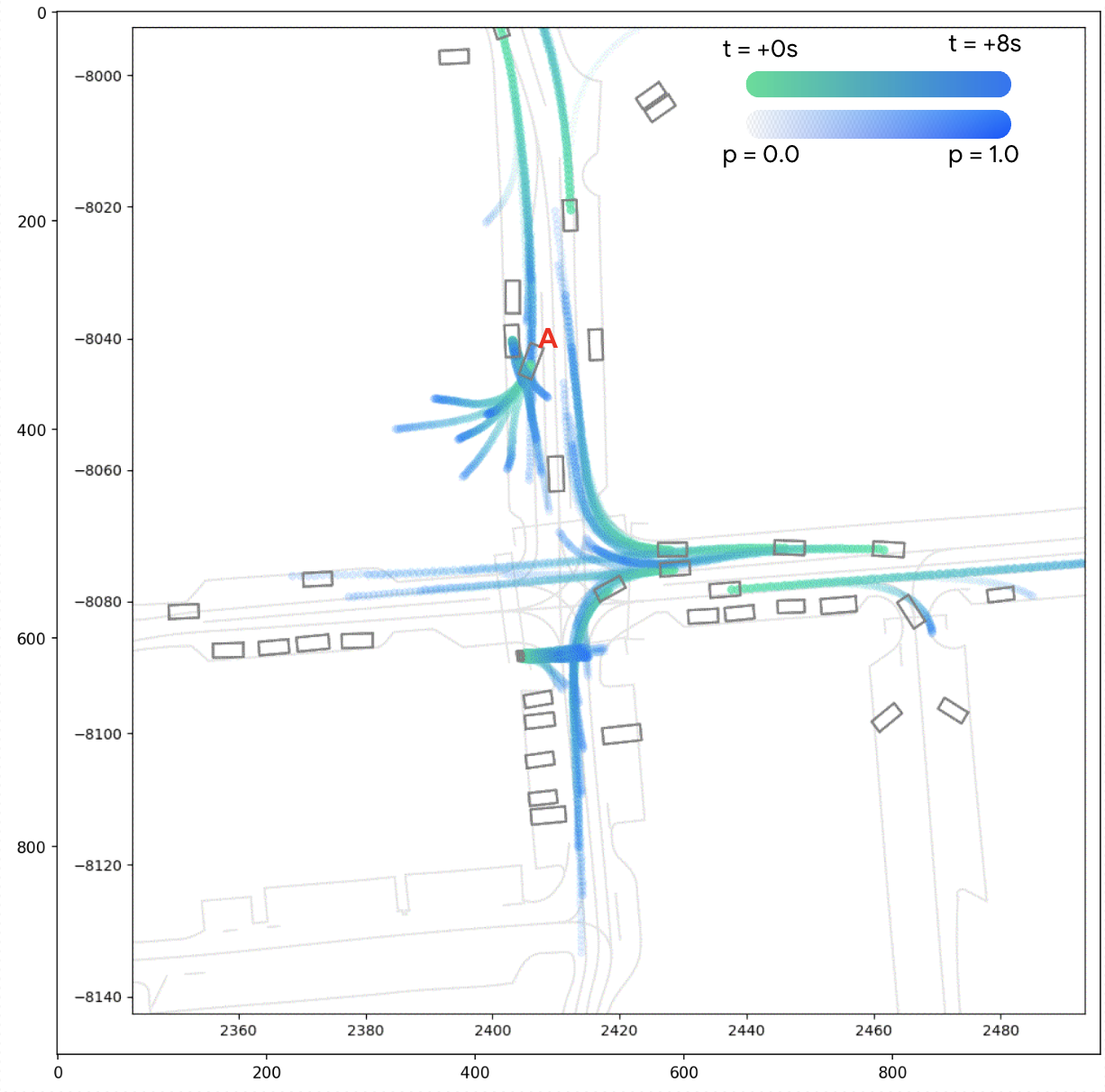}
    \caption{MultiPath++ (MP++)}
  \end{subfigure}
  \begin{subfigure}[b]{0.49\columnwidth}
    \includegraphics[width=\linewidth]{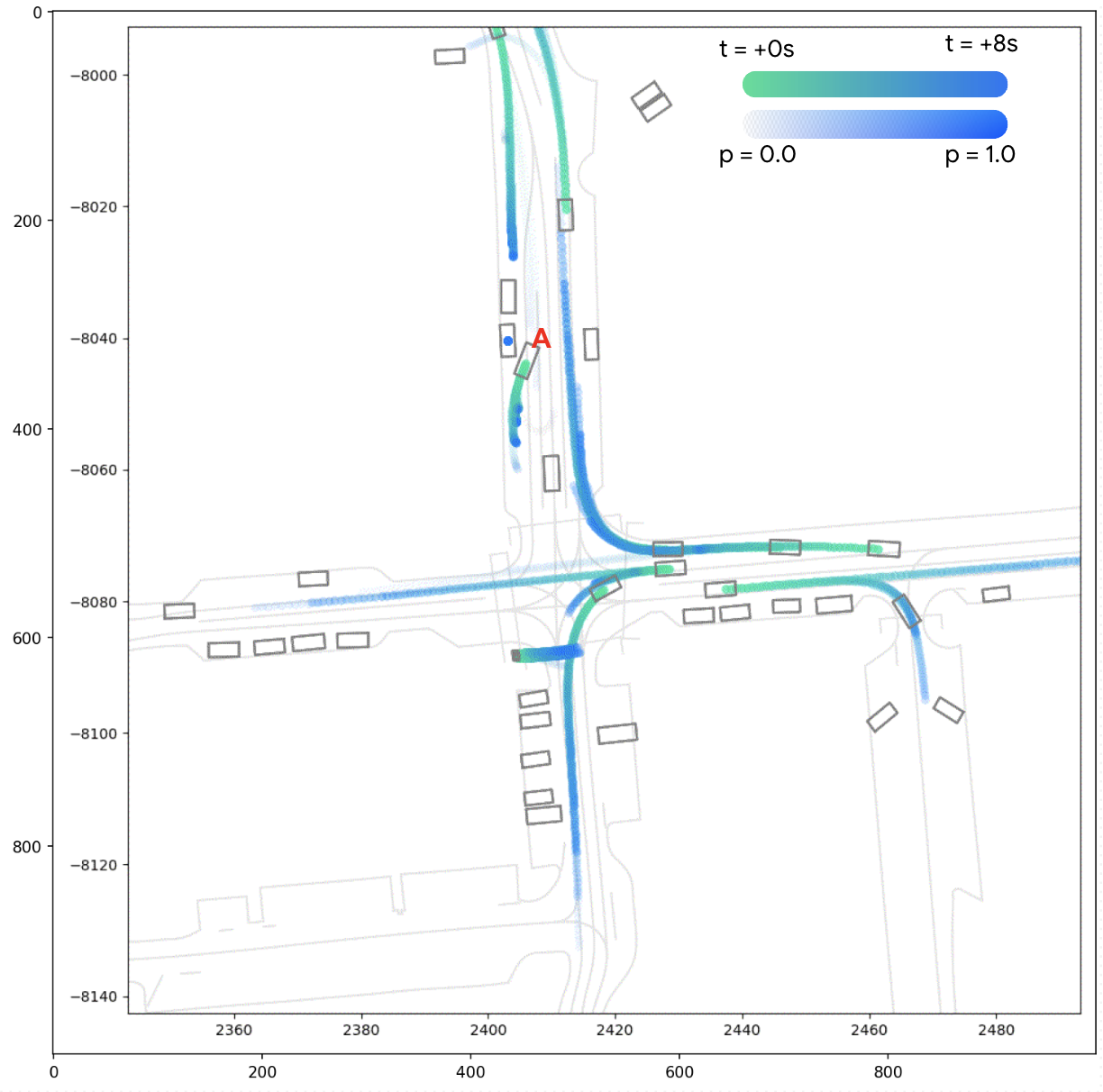}
    \caption{Wayformer (WF)}
  \end{subfigure}
  \caption{In this scenario we observe that MP++ is not able to model the future of the vehicle (agent A) entering the parking lane and outputs a multi-modal equally likely future for this agent. WF understands the roadgraph interaction much better and outputs trajectories that have high likelihood that agent A is entering the parking lane.}
  \label{fig:qualitative-wins-5}
\end{figure}

\begin{figure}[h]
  \begin{subfigure}[b]{0.49\columnwidth}
    \includegraphics[width=\linewidth]{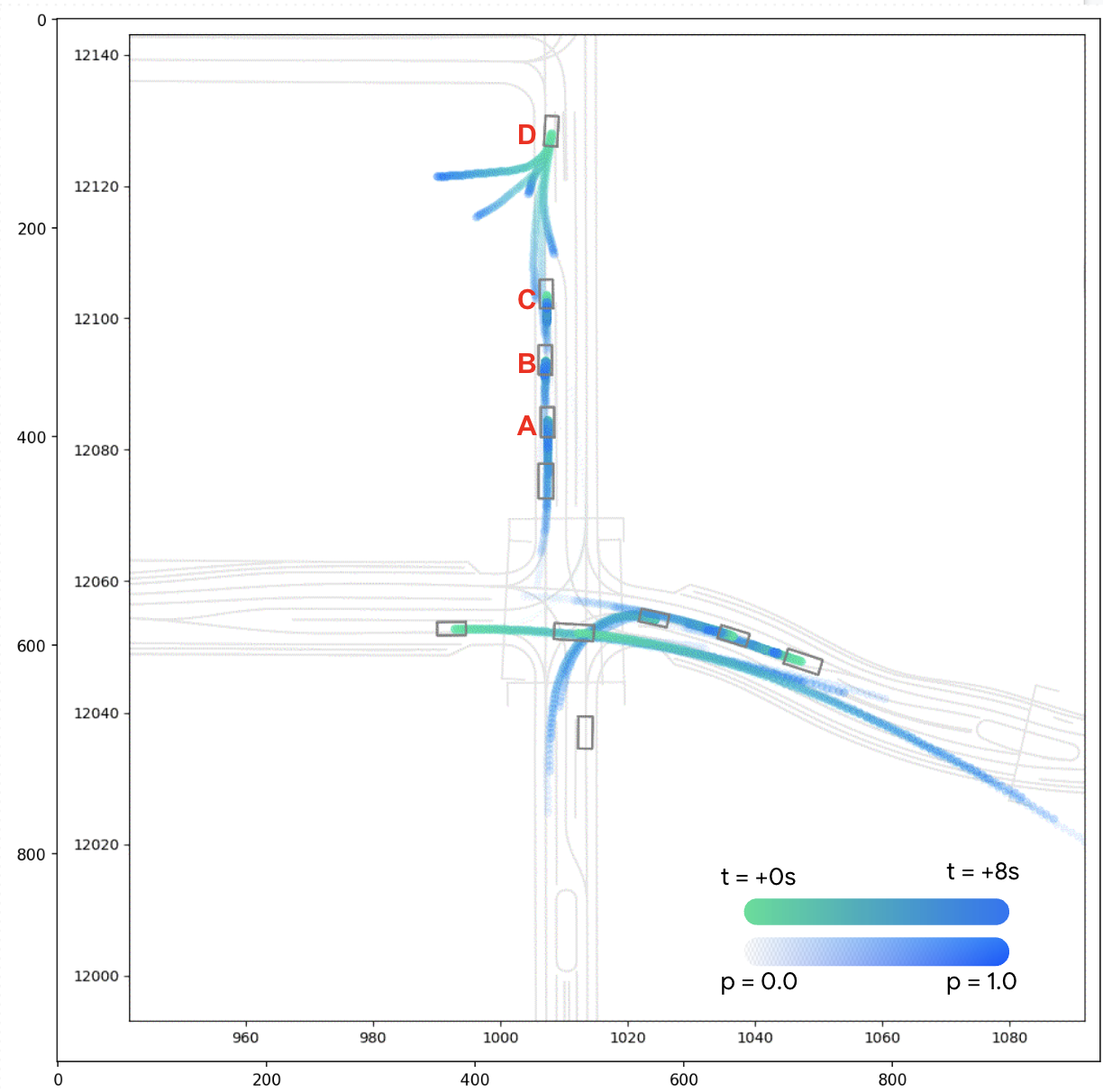}
    \caption{MultiPath++ (MP++)}
  \end{subfigure}
  \begin{subfigure}[b]{0.49\columnwidth}
    \includegraphics[width=\linewidth]{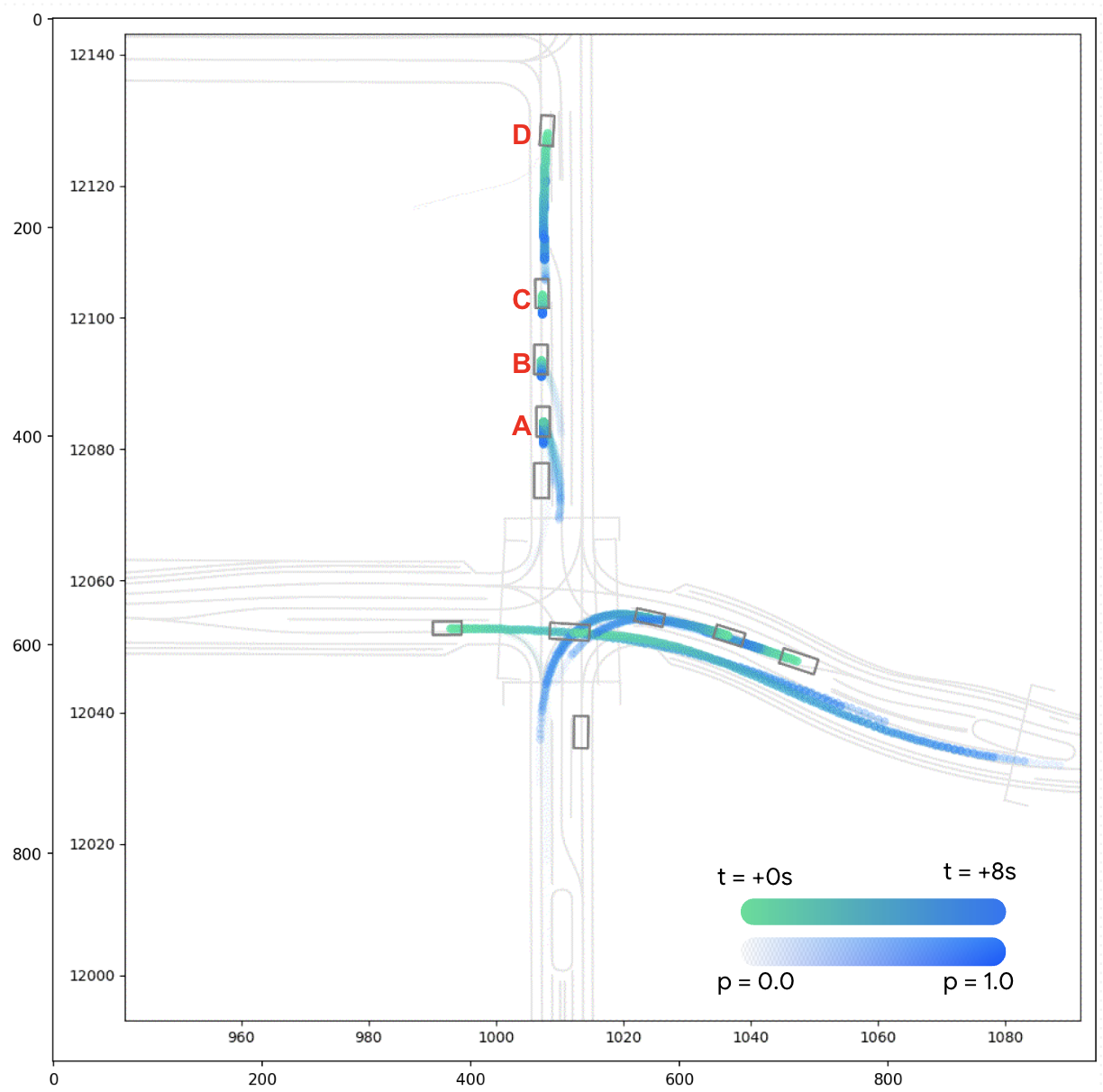}
    \caption{Wayformer (WF)}
  \end{subfigure}
  \caption{We see agents A, B, and C are waiting behind a stationary vehicle. WF predicts agent A will nudge around the stationary vehicle to make progress, while MP predicts the agents will proceed through the stationary vehicle. Additionally, MP predicts agent D could proceed off the road, while WF predicts it to follow the road behind agent C.}
  \label{fig:qualitative-wins-6}
\end{figure}

\begin{figure}[h]
  \begin{subfigure}[b]{0.49\columnwidth}
    \includegraphics[width=\linewidth]{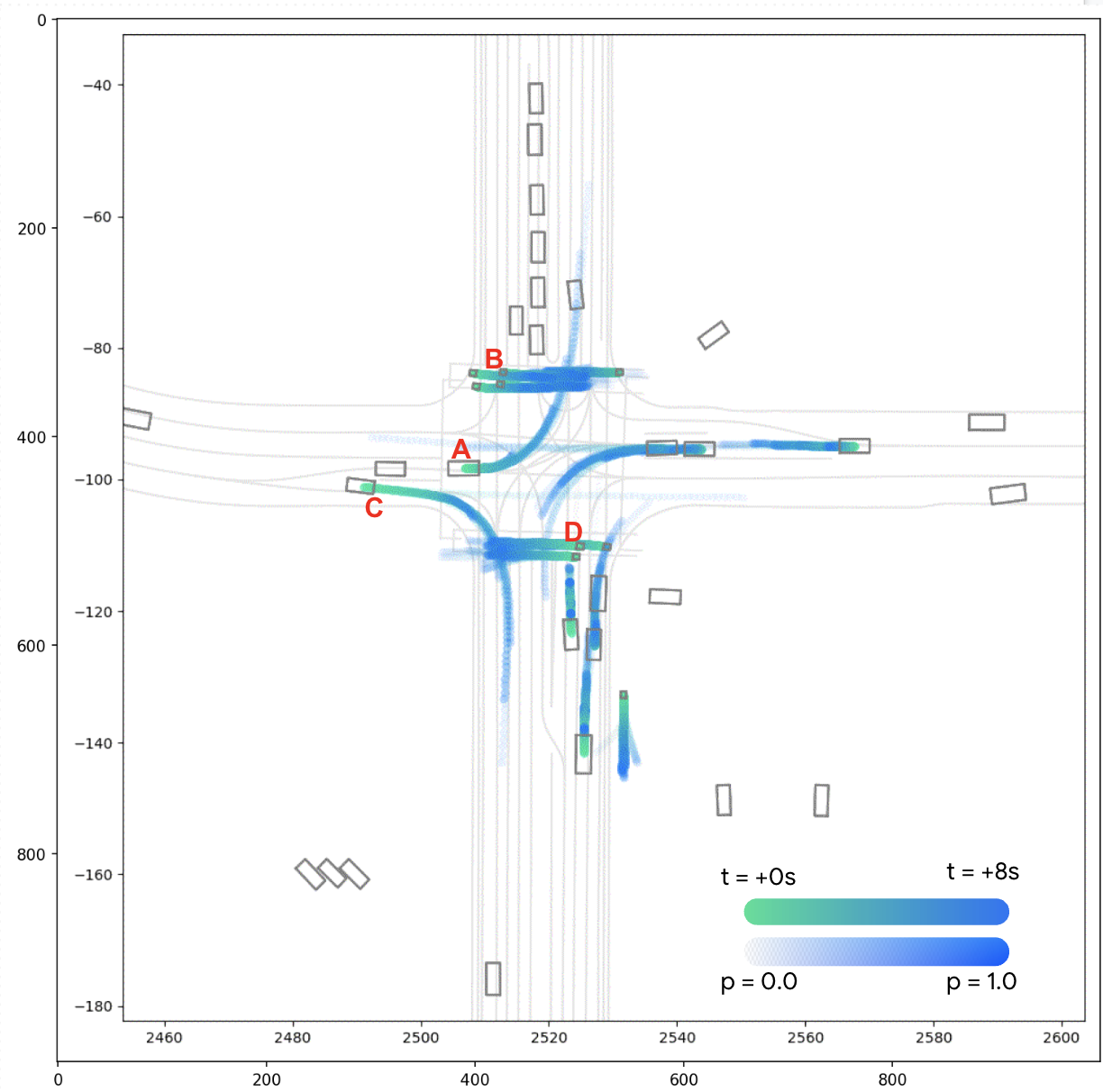}
    \caption{MultiPath++ (MP++)}
  \end{subfigure}
  \begin{subfigure}[b]{0.49\columnwidth}
    \includegraphics[width=\linewidth]{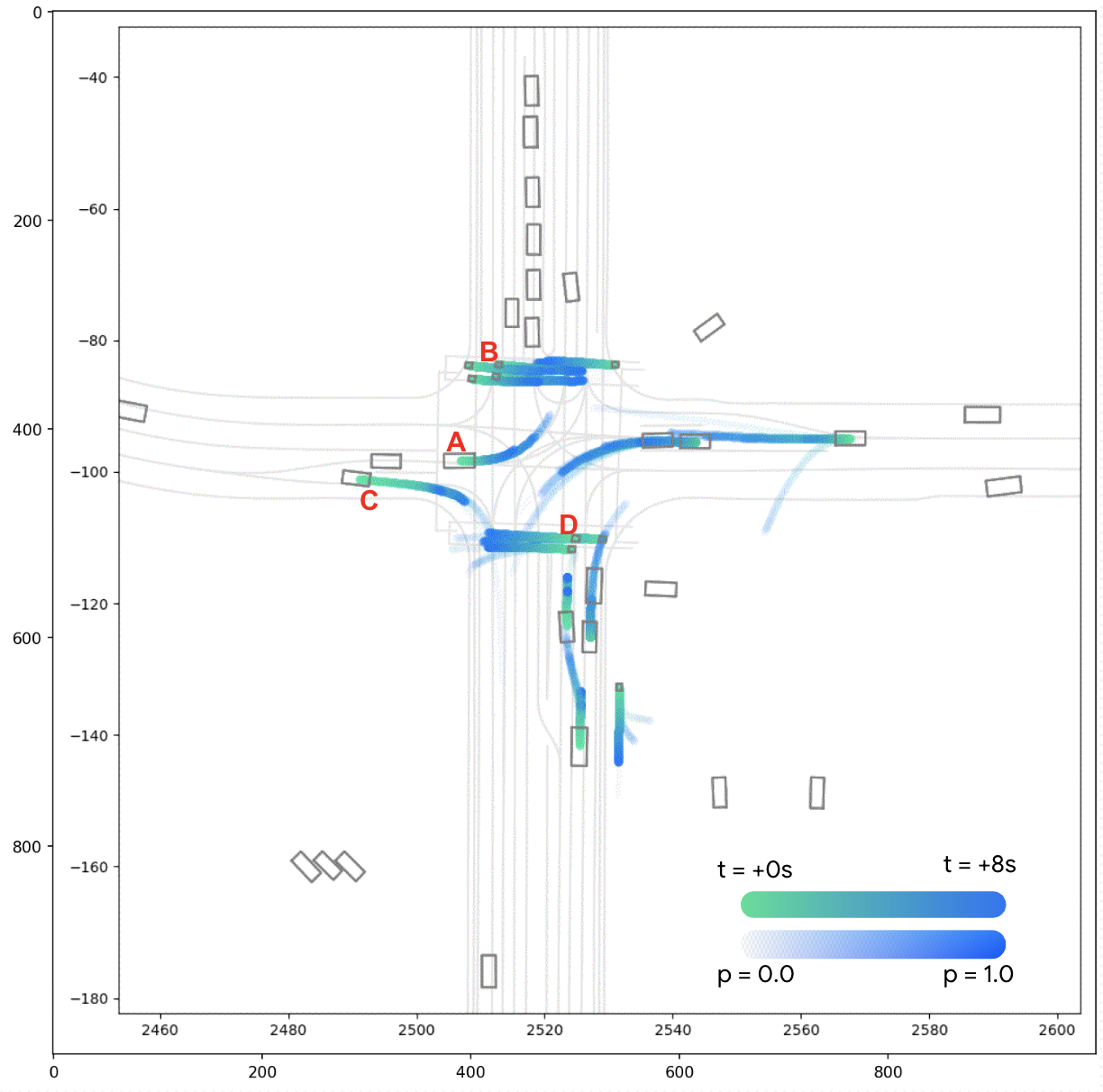}
    \caption{Wayformer (WF)}
  \end{subfigure}
  \caption{Multiple pedestrians, including agent B, are crossing the road and both MP++ and WF predict car A wants to make a left turn through that crosswalk. WF predicts car A will start to turn, then wait as the pedestrians cross, while MP++ predicts that car A will proceed through the crosswalk even as the pedestrians are crossing.}
  \label{fig:qualitative-wins-7}
\end{figure}

\begin{figure}[h]
  \begin{subfigure}[b]{0.49\columnwidth}
    \includegraphics[width=\linewidth]{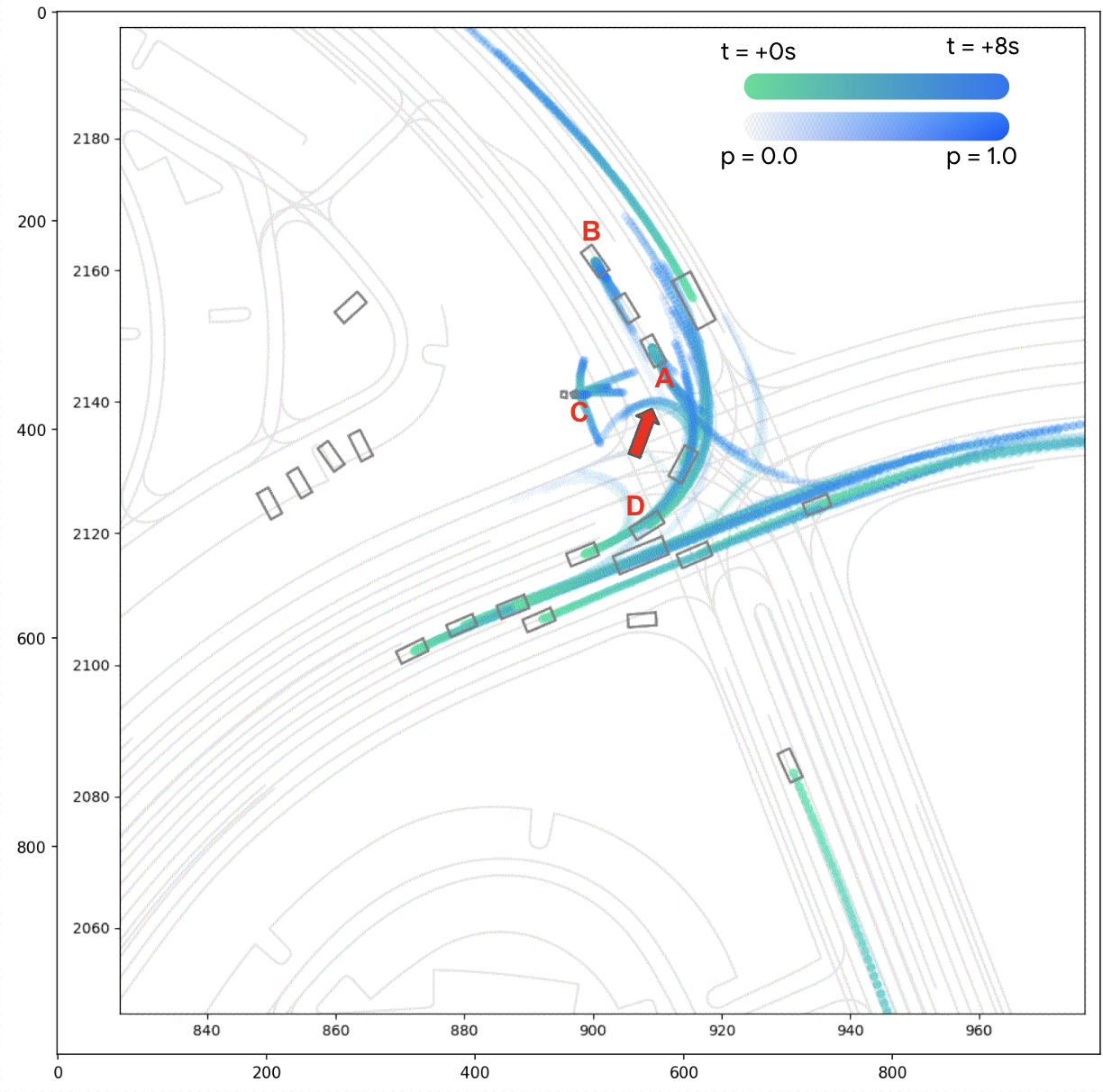}
    \caption{MultiPath++ (MP++)}
  \end{subfigure}
  \begin{subfigure}[b]{0.49\columnwidth}
    \includegraphics[width=\linewidth]{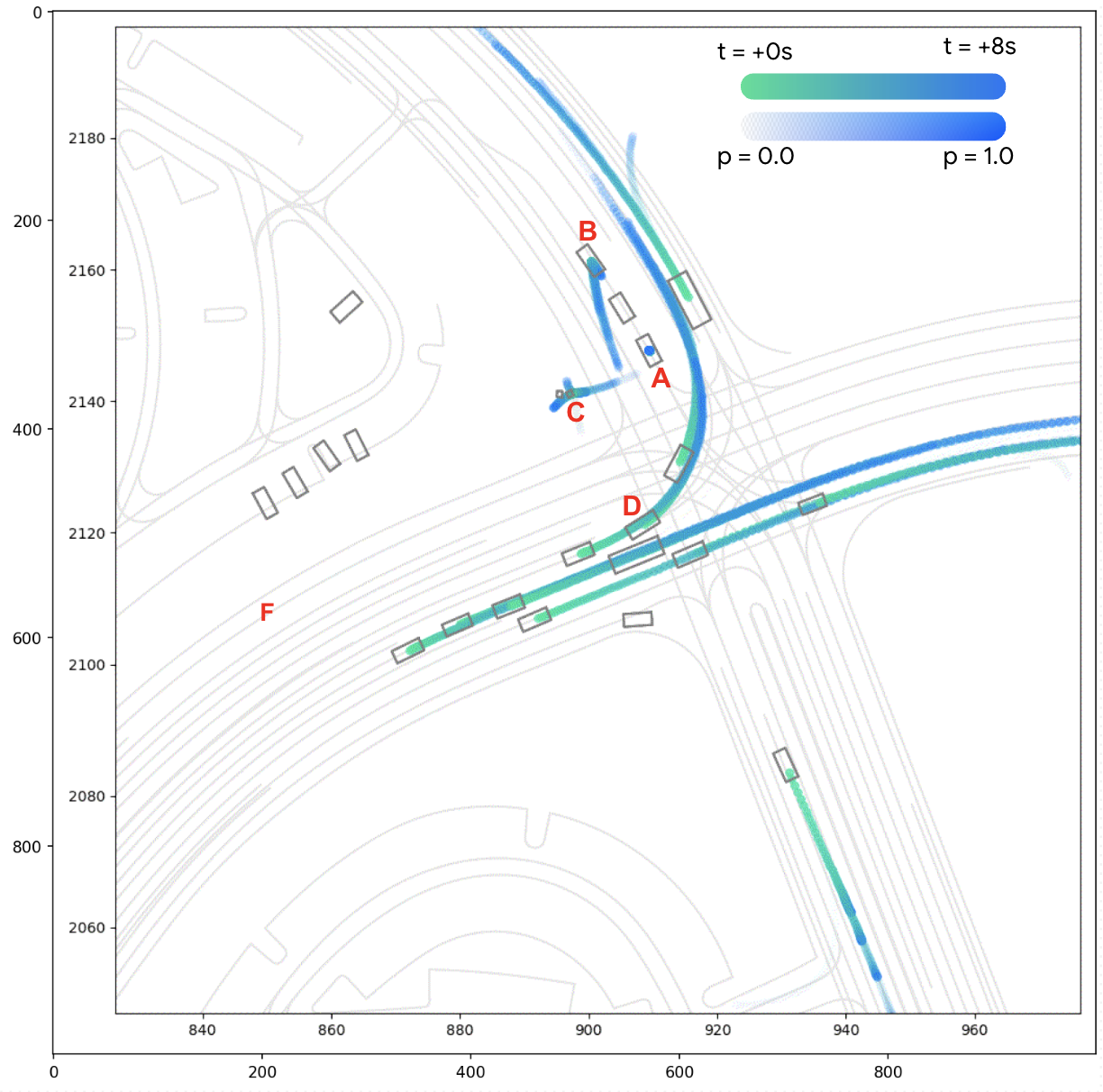}
    \caption{Wayformer (WF)}
  \end{subfigure}
  \caption{This shows a busy intersection, with both WF and MP++ predicting vehicles in the left-right road (i.e. agent D) are either proceeding straight or left turning. However, MP++ predicts agent A to try to make a left turn directly into the flow of traffic, including through other cars left turning, while WF predicts agent A will wait. Additionally, MP++ predicts agent B will try to proceed through the vehicle waiting in front of it, while WF instead predicts it either remaining stationary or nudging to the adjacent lane. Furthermore, WF also predicts agent D to potentially make a U-turn that goes through the corner of the sidewalk near agent C (highlighted by the red arrow).}
  \label{fig:qualitative-wins-8}
\end{figure}



\begin{figure}[h]
  \begin{subfigure}[b]{0.49\columnwidth}
    \includegraphics[width=\linewidth]{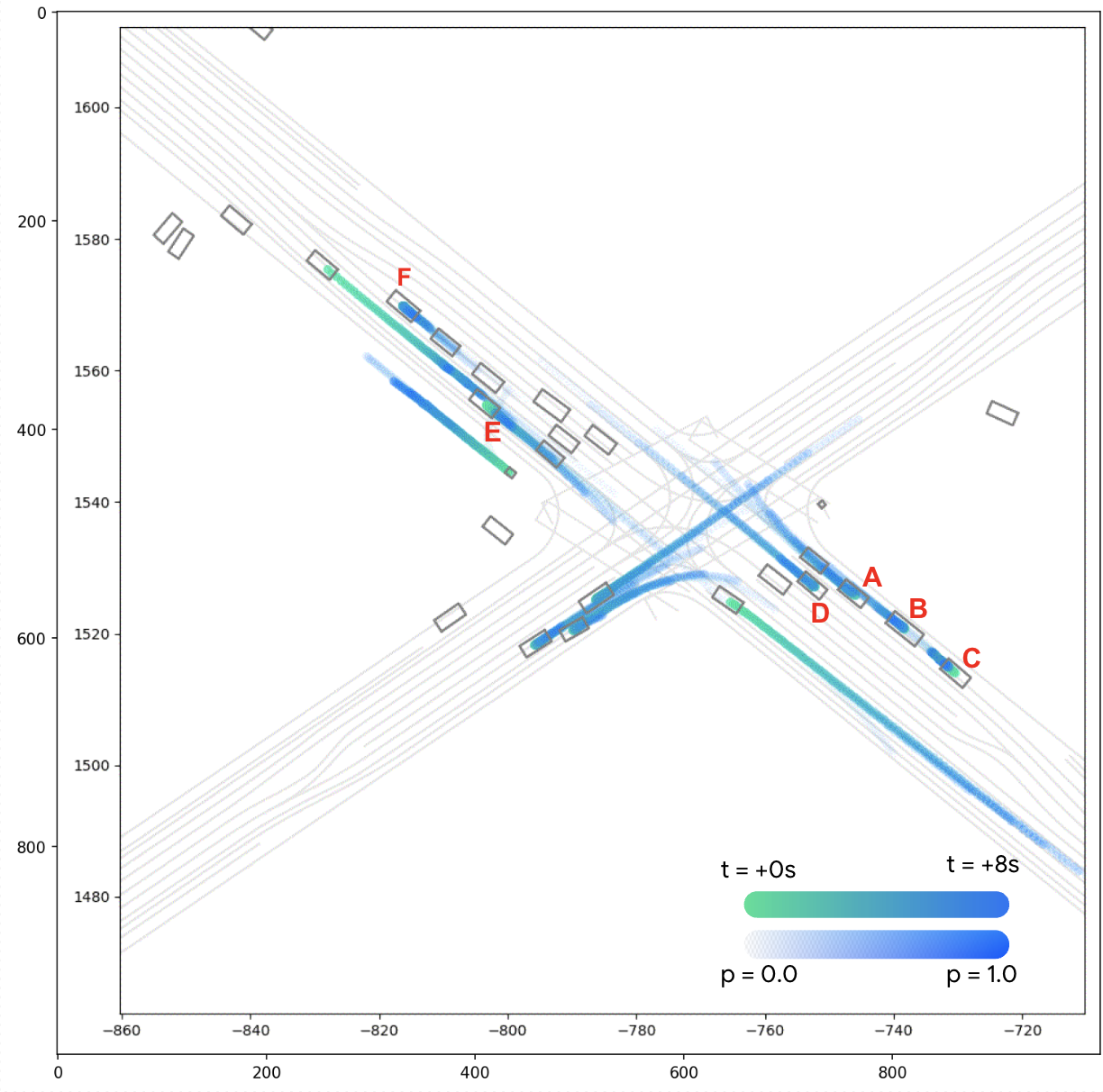}
    \caption{MultiPath++ (MP++)}
  \end{subfigure}
  \begin{subfigure}[b]{0.49\columnwidth}
    \includegraphics[width=\linewidth]{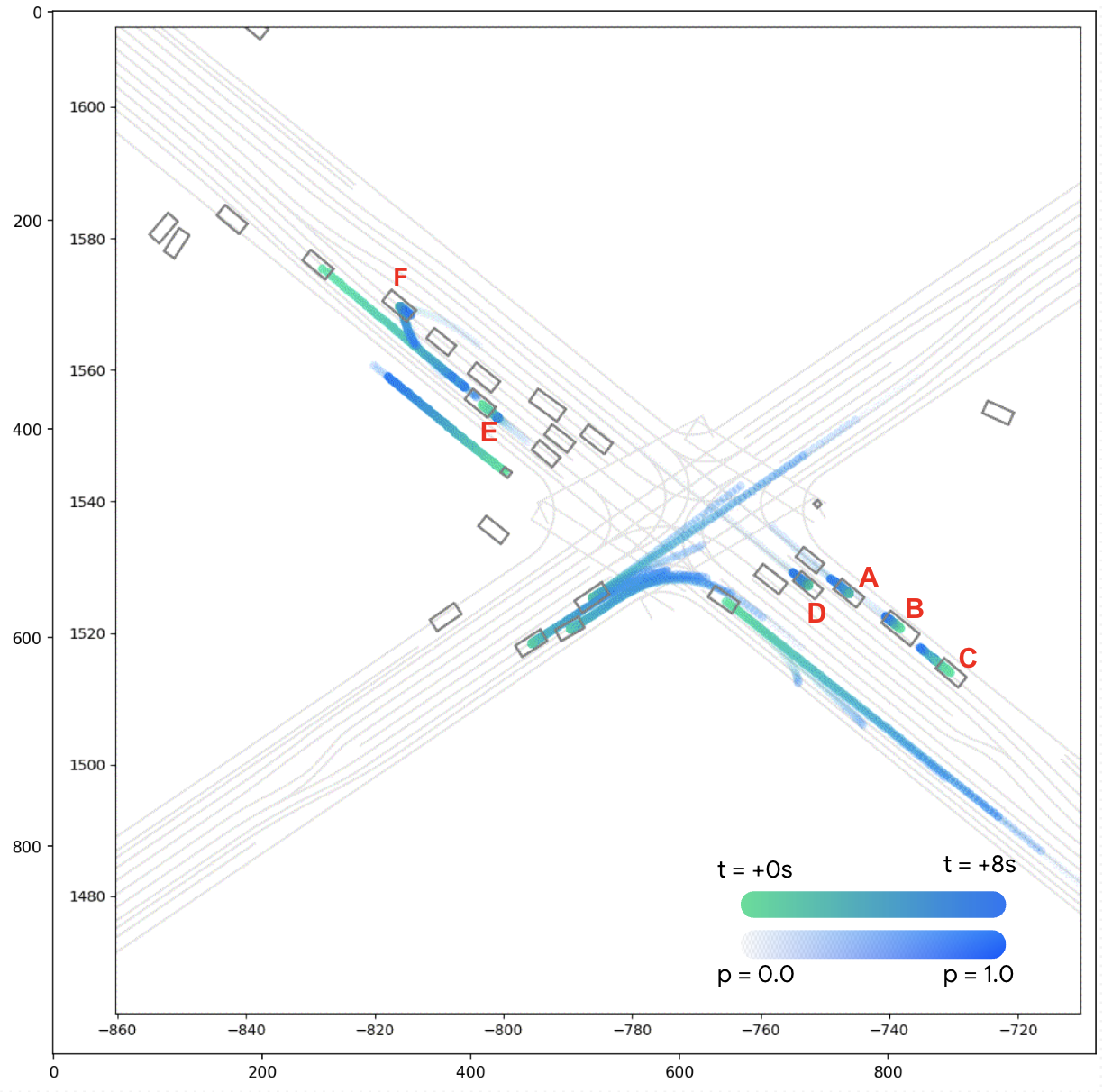}
    \caption{Wayformer (WF)}
  \end{subfigure}
  \caption{This scenario represents a 4-way intersection. (a) At the intersection vehicles A, B, C and D are all stopped at the intersection due to signal. WF takes this into account and predicts yielding behavior. Vehicle D yielding for the light, Vehicle C yielding for B, Vehicle B yielding for A and Vehicle A yielding for the vehicle in-front. But, MP++'s predictions for the same agents go through the intersection (Vehicle D) and for vehicles A, B and C, they pass through the vehicles in-front. (b) We see similar behaivor on the other side of the intersection, where vehicle E's WF predictions are yielding and MP++ predictions are passing through vehicles in the front.}
  \label{fig:qualitative-wins-11}
\end{figure}

\begin{figure}[h]
  \begin{subfigure}[b]{0.49\columnwidth}
    \includegraphics[width=\linewidth]{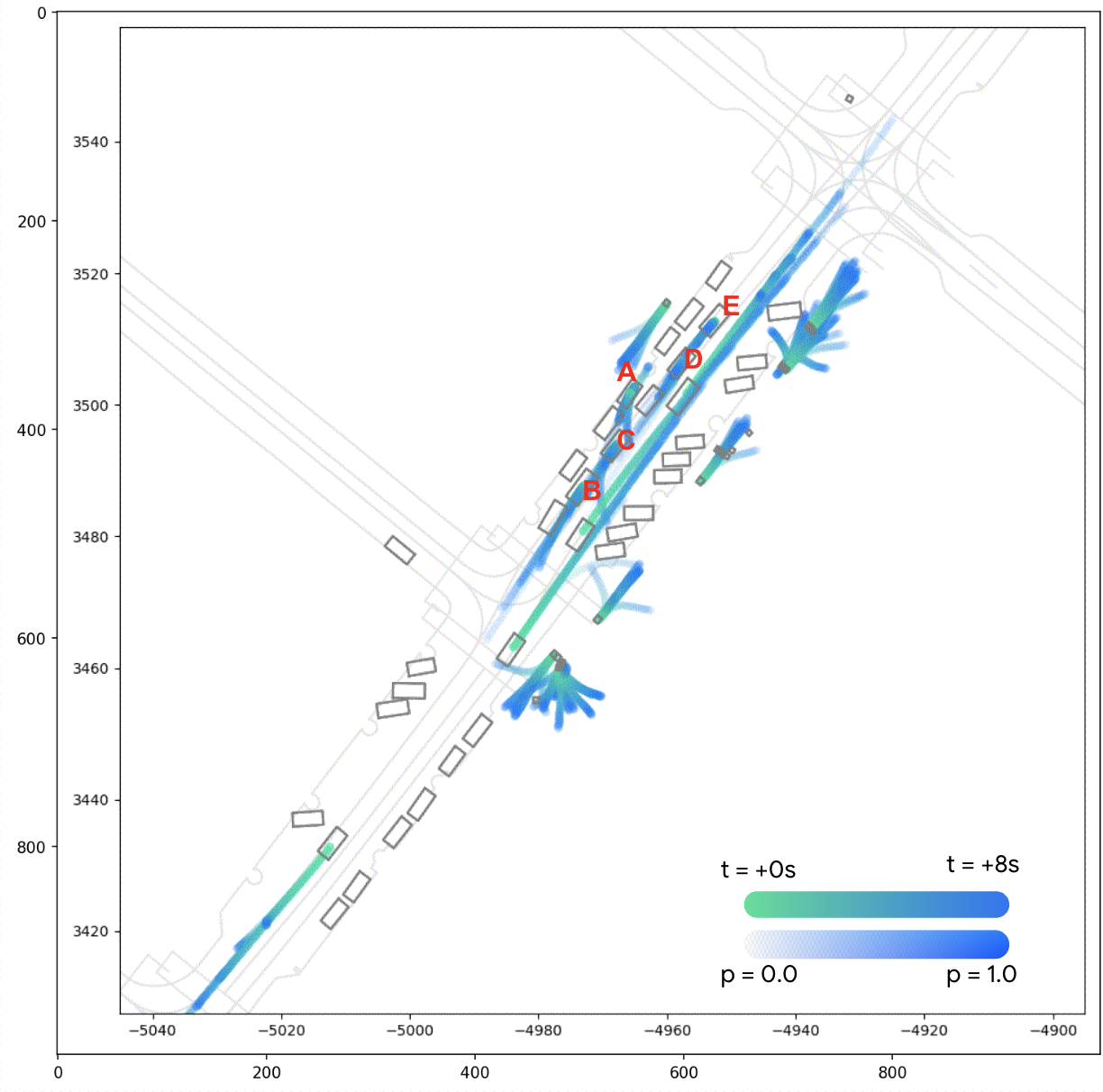}
    \caption{MultiPath++ (MP++)}
  \end{subfigure}
  \begin{subfigure}[b]{0.49\columnwidth}
    \includegraphics[width=\linewidth]{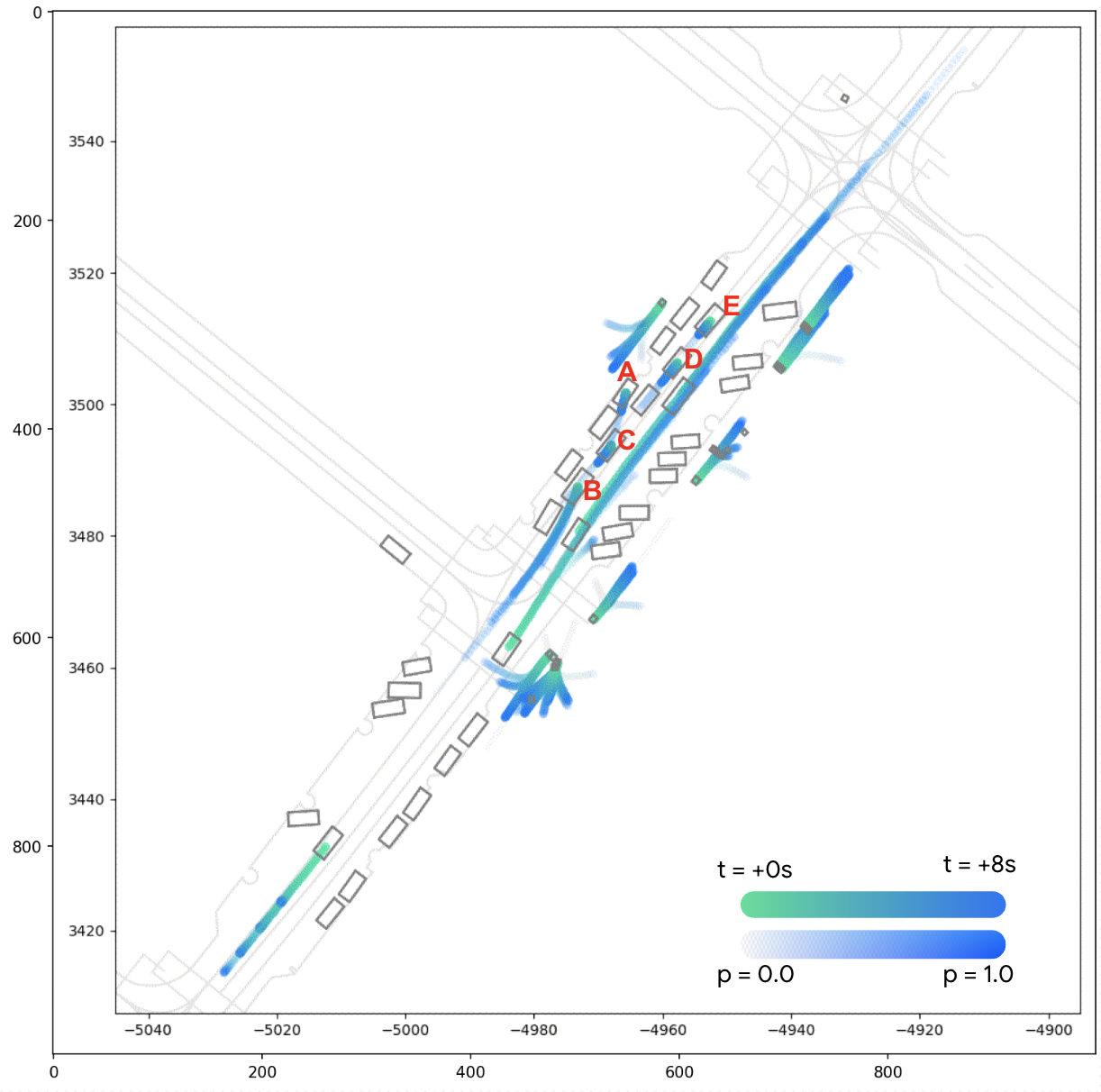}
    \caption{Wayformer (WF)}
  \end{subfigure}
  \caption{This scenario represents a T intersection with narrow roads and parked cars. In this highly interactive scene, we observe that (a) a parked vehicle (vehicle A) is trying to merge into traffic. WF predicts nudging around already parked cars and merging onto the traffic, while MP++ predictions pass through the parked cars in front of A. (b) In addition, we also see that for vehicle B WF predicts that nudges around the vehicle in front while MP++ predictions go through the car in-front. (c) For vehicles C, D and E, WF predicts yielding behavior (C yielding for B, D yielding for car in the front and E yielding for D), while MP++ predictions go through the vehicles in front.}
  \label{fig:qualitative-wins-12}
\end{figure}

\begin{figure}[h]
  \begin{subfigure}[b]{0.49\columnwidth}
    \includegraphics[width=\linewidth]{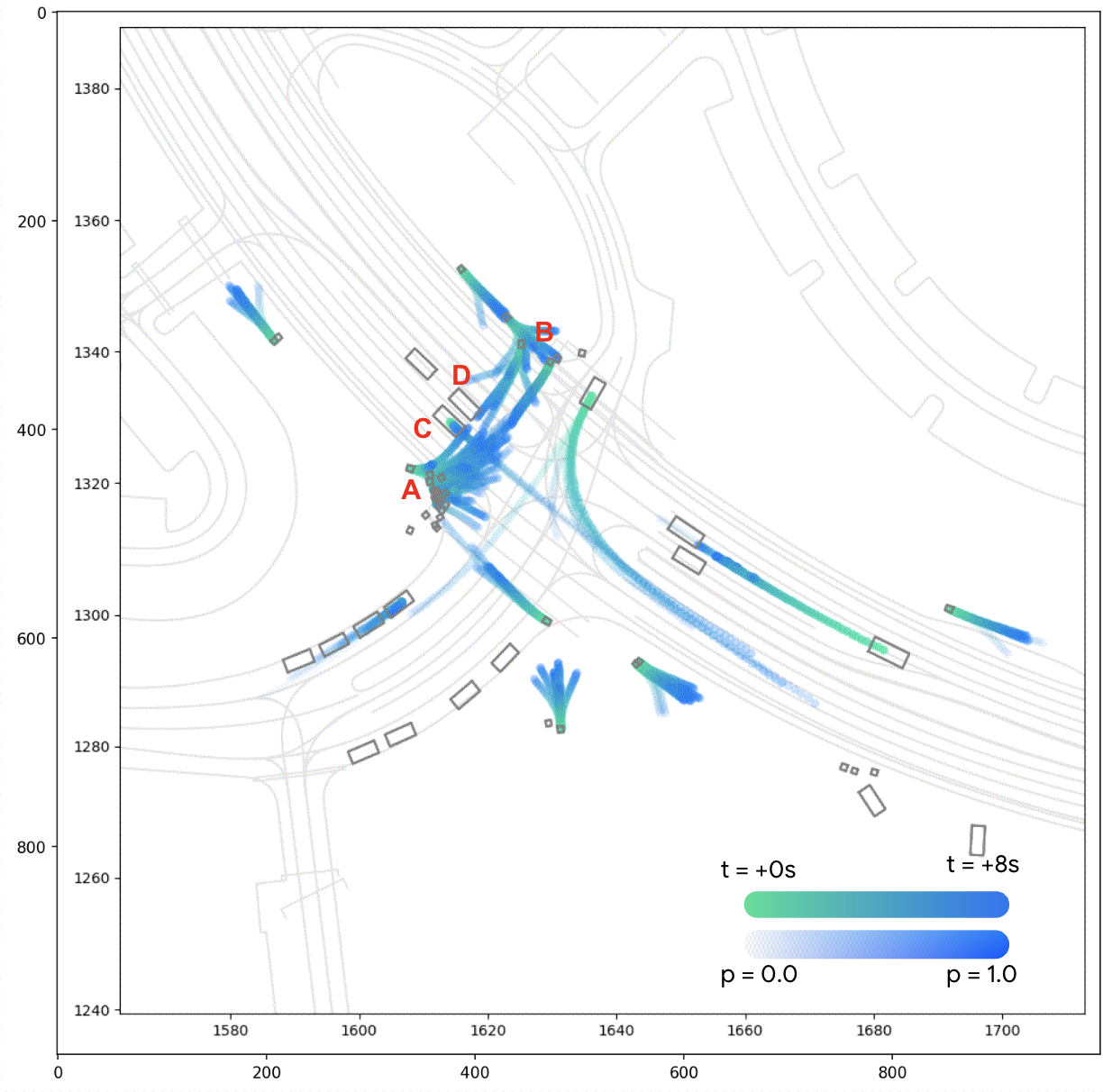}
    \caption{MultiPath++ (MP++)}
  \end{subfigure}
  \begin{subfigure}[b]{0.49\columnwidth}
    \includegraphics[width=\linewidth]{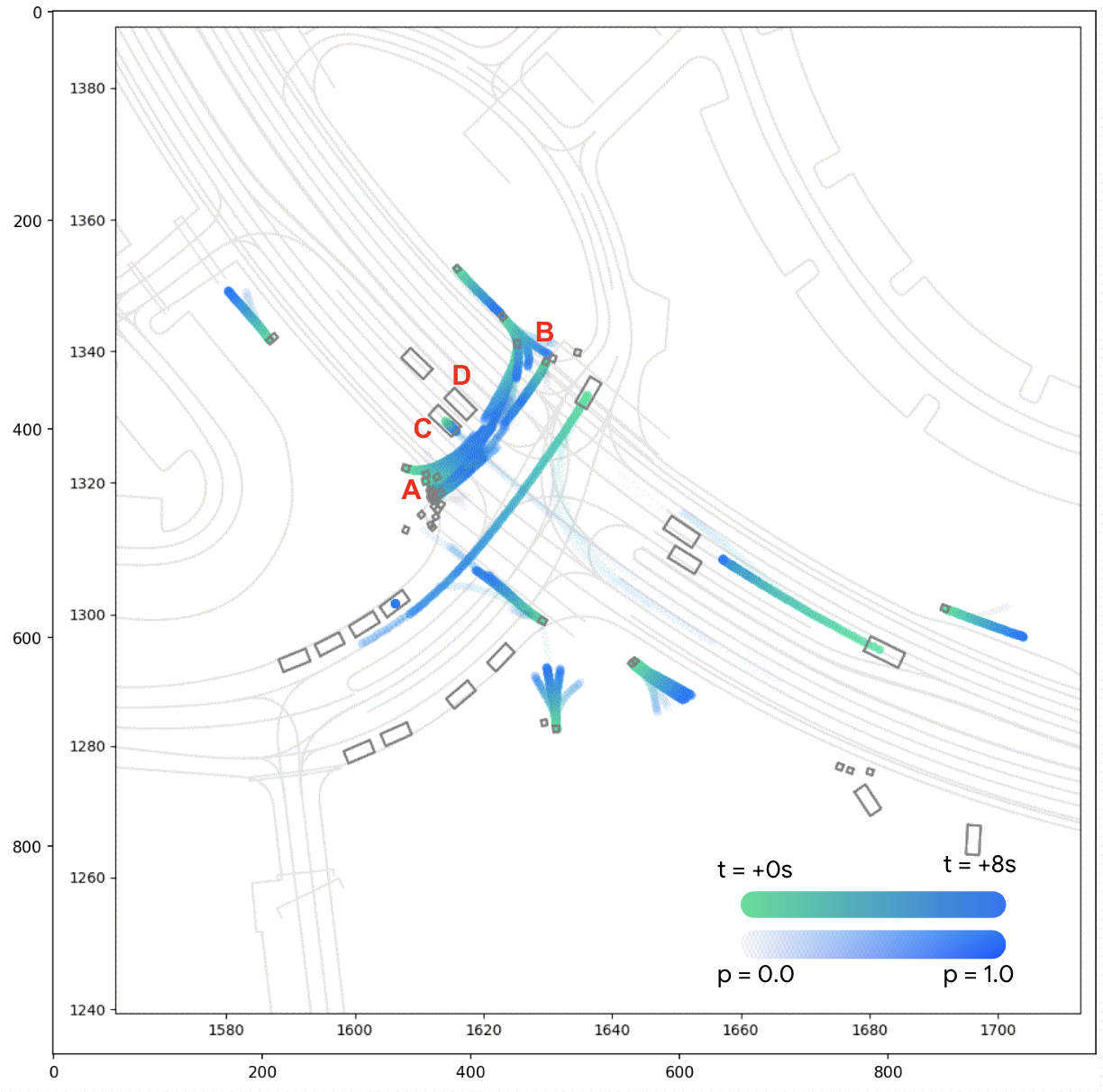}
    \caption{Wayformer (WF)}
  \end{subfigure}
  \caption{This scenario represents a very busy 4-way intersection with clusters of pedestrians (A, B). Both these clusters are pedestrians  crossing the signal from either side of the road. We observe that MP++ prediction's are more distributed, some of them going through already stopped vehicles (vehicles C and D) at the intersection. But, WF understands the presence of other vehicles and produces predictions which do not cross through them. We also see that WF's predictions for vehicle C yield to pedestrians while MP++'s predictions do not.}
  \label{fig:qualitative-wins-13}
\end{figure}
\end{document}